# Representing and Reasoning with Qualitative Preferences for Compositional Systems

**Ganesh Ram Santhanam**                                        GSANTHAN@CS.IASTATE.EDU
**Samik Basu**                                                        SBASU@CS.IASTATE.EDU
**Vasant Honavar**                                                 HONAVAR@CS.IASTATE.EDU
*Department of Computer Science*
*Iowa State University*
*Ames, IA 50011, USA*

## Abstract

Many applications, e.g., Web service composition, complex system design, team formation, etc., rely on methods for identifying collections of objects or entities satisfying some functional requirement. Among the collections that satisfy the functional requirement, it is often necessary to identify one or more collections that are optimal with respect to user preferences over a set of attributes that describe the non-functional properties of the collection.

We develop a formalism that lets users express the relative importance among attributes and qualitative preferences over the valuations of each attribute. We define a *dominance* relation that allows us to compare collections of objects in terms of preferences over attributes of the objects that make up the collection. We establish some key properties of the dominance relation. In particular, we show that the dominance relation is a strict partial order when the intra-attribute preference relations are strict partial orders and the relative importance preference relation is an interval order.

We provide algorithms that use this dominance relation to identify the set of most preferred collections. We show that under certain conditions, the algorithms are guaranteed to return *only* (sound), *all* (complete), or *at least one* (weakly complete) of the most preferred collections. We present results of simulation experiments comparing the proposed algorithms with respect to (a) the quality of solutions (number of most preferred solutions) produced by the algorithms, and (b) their performance and efficiency. We also explore some interesting conjectures suggested by the results of our experiments that relate the properties of the user preferences, the dominance relation, and the algorithms.

## 1. Introduction

Many applications call for techniques for representing and reasoning about preferences over a set of alternatives. In such settings, preferences over the alternatives are expressed with respect to a set of attributes that describe the alternatives. Such preferences can be either *qualitative* or *quantitative*. A great deal of work on multi-attribute decision theory has focused on reasoning with quantitative preferences (Fishburn, 1970a; Keeney & Raiffa, 1993). However, in many settings it is more natural to express preferences in qualitative terms (Doyle & Thomason, 1999) and hence, there is a growing interest on formalisms for representing and reasoning with qualitative preferences (Brafman & Domshlak, 2009) in AI.

An important problem in this context has to do with representing qualitative preferences over multiple attributes and reasoning with them to find the most preferred among a set of alternatives. Brafman, Domshlak and Shimony's seminal work (2006) attempts to address this problem by introducing *preference networks* that capture: (a) *intra-variable or intra-attribute* preferences specifying preferences over the domains of attributes; (b) the *relative importance* among the attributes. Preference networks use a graphical representation to compactly encode the above types of preferences from the user, and employ the *ceteris paribus*[1] semantics to reason about the most preferred alternatives. In this model, each alternative is completely described by the values assigned to a set of attributes.

In many AI applications such as planning and scheduling, the alternatives have a *composite* structure, i.e., an alternative represents a collection or a *composition* of objects rather than simple objects. In such settings, typically there are a set of user specified functional requirements that compositions are required to satisfy[2]. Among all the possible compositions that do satisfy the functional requirements, there is often a need to choose compositions that are most preferred with respect to a set of user preferences over a set of *non-functional* attributes of the objects that make up the composition. We illustrate the above problem using the following example.

## 1.1 Illustrative Example

Consider the task of designing a program of study (POS) for a Masters student in the Computer Science department. The POS consists of a collection of courses chosen from a given repository of available courses spanning different areas of focus in computer science. Apart from the area of focus, each course also has an assigned instructor and a number of credit hours. A repository of available courses, their areas of focus, their instructors and the number of credit hours are specified in Table 1.

| Course | Area | Instructor | Credits |
|--------|------|------------|---------|
| CS501 | Formal Methods (FM) | Tom | 4 |
| CS502 | Artificial Intelligence (AI) | Gopal | 3 |
| CS503 | Formal Methods (FM) | Harry | 2 |
| CS504 | Artificial Intelligence (AI) | White | 3 |
| CS505 | Databases (DB) | Bob | 4 |
| CS506 | Networks (NW) | Bob | 2 |
| CS507 | Computer Architecture (CA) | White | 3 |
| CS508 | Software Engineering (SE) | Tom | 2 |
| CS509 | Theory (TH) | Jane | 3 |
| CS510 | Theory (TH) | Tom | 3 |

Table 1: List of courses the student can choose from

In this example, each POS can be viewed as a composition of courses. The requirements for an acceptable Masters POS (i.e., a feasible composition) are as follows.

$F$1. The POS should include at least 15 credits

---

1. A Latin term for 'all else being equal'
2. For example, in planning, a valid plan is a collection of actions that satisfies the goal; and in scheduling, a valid schedule is a collection of task-to-resource assignments that respects the precedence constraints.

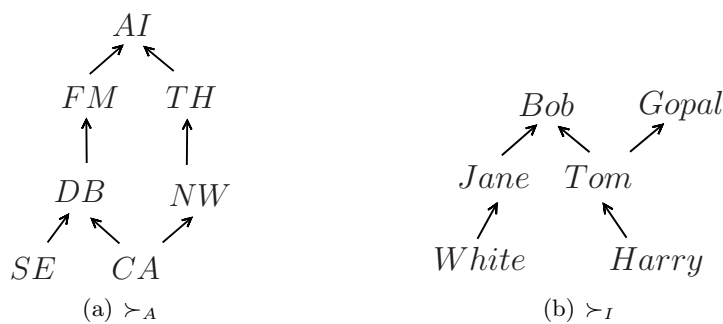

Figure 1: Intra-attribute preferences for Area ($\succ_A$) and Instructor ($\succ_I$).

$F2.$ The POS should include the two core courses CS509 and CS510

$F3.$ There should be courses covering at least two breadth areas of study (apart from the area of Theory (TH))

Given the repository of courses (see Table 1; there may be one or more acceptable programs of study, i.e., feasible compositions). For example:

- $P_1 = CS501 \oplus CS502 \oplus CS503 \oplus CS504 \oplus CS509 \oplus CS510$

- $P_2 = CS501 \oplus CS502 \oplus CS505 \oplus CS506 \oplus CS509 \oplus CS510$

- $P_3 = CS503 \oplus CS504 \oplus CS507 \oplus CS508 \oplus CS509 \oplus CS510$

Suppose that in addition to the above requirements, a student has some preferences over the course attributes such as the area of focus, the choice of instructors and difficulty level in terms of credit hours. Among several acceptable programs of study, the student may be interested in those programs of study that: (a) satisfy the minimum requirements (see above) for an acceptable POS, and (b) those that are most preferred with respect to his/her preferences specified above. The preferences of a student with respect to the course attributes Area ($A$) and Instructor ($I$) are illustrated in Figure 1 (arrows are directed toward the preferred area/instructor in the figure, e.g., $AI$ is preferred to $FM$ and $Bob$ is preferred to $Tom$). In addition let us say that the student prefers the POS that have lesser total number of credits (this specifies $\succ_C$). Further, let the relative importance among the attributes $A$, $I$ and $C$ be $I \rhd A \rhd C$, i.e., $I$ is relatively more important than $A$, which is in turn relatively more important than $C$.

## 1.2 Problem Statement for the Illustrative Example

The problems that we try to address in this paper for the above example are:

- Given two programs of study, namely $P_i$ and $P_j$, determine whether $P_i$ *dominates* (i.e., is preferred to) $P_j$ or vice versa with respect to the student's preferences;

- Given a repository of courses and an algorithm for computing a set of acceptable programs of study, find the most preferred, acceptable programs of study with respect to the above dominance relation.

In the example given in Section 1.1, the functional requirements correspond to the three conditions $F1$ to $F3$, all of which must be satisfied for a collection of courses to be an acceptable POS. Area $(A)$, instructor $(I)$ and number of credits $(C)$ constitute the non-functional attributes, and the user preferences over these attributes are given by $\{\succ_A, \succ_I, \succ_C\}$ and $I \rhd A \rhd C$. One can envision similar problems in several other applications, ranging from assembling hardware and software components in an embedded system (such as designing a pacemaker or anti-lock braking system) to putting together a complex piece of legislation (such as the one for reforming health care).

In general, we are interested in the problem of (a) reasoning about preferences over compositions of objects, given the preferences over a set of non-functional attributes describing the objects; and (b) identifying compositions that satisfy the functional requirements of the compositional system, and at the same time are optimal with respect to the stated preferences over the non-functional attributes. Against this background, we present a preference formalism and a set of algorithms to address this problem in compositional systems.

## 1.3 Contributions

We adopt the preference network representation introduced by Brafman et al. (2006) for the specification of qualitative preferences[3] over valuations of each attribute as well as the relative importance among the attributes. We extend reasoning about preferences over single objects to deal with preferences over collections of objects. The main contributions of this paper are as follows.

1. We develop a preference formalism that allows users to specify preferences in terms of intra-attribute and relative importance preferences over a set of attributes, and includes mechanisms for:

   a) Computing the valuation of a composition: With respect to each attribute, we define a generic *aggregation function* to compute the valuation of a composition as a function of the valuations of its components. We also present a strict partial order preference relation for comparing two compositions with respect to their aggregated valuations of each attribute.

   b) Comparing the valuations of compositions: We introduce a *dominance* relation that compares compositions (in terms of their aggregated valuations) with respect to the stated preferences, and establish some of its key properties. In particular, we show that this relation is a strict partial order whenever the intra-attribute preferences are strict partial orders and relative importance preference is an interval order.

2. We develop a suite of algorithms that identify the set, or subset of the most preferred composition(s) with respect to the user preferences. In particular, we show that under certain conditions, the algorithms are guaranteed to return only (sound), all (complete), or at least one (weakly complete) of the most preferred compositions. The algorithms we develop fall into two classes:

---

3. We do not deal with conditional preferences in this work.

a) those that first compute the set of all feasible compositions using a functional composition algorithm as a black box, and then proceed to find the most preferred among them using the preference relations developed in (1); and

b) an algorithm that *interleaves* at each step the execution of a functional composition algorithm and the ordering of partial solutions with respect to user preferences. It requires the functional composition algorithm to be able to construct a composition satisfying the functional requirement incrementally, i.e., by iteratively extending partial compositions with additional components.

We analyze some key properties of the algorithms that yield specific conditions on the structure of preferences, under which the algorithms produce only/at least one/all of the most preferred solutions.

3. We present results of experiments that compare performance of the above algorithms for computing the most preferred compositions on a set of simulated composition problem instances. The results demonstrate the feasibility of our approach in practice, and compare our algorithms with respect to the quality of (number of *good* or most preferred) solutions produced by the algorithms and their performance (running time) and efficiency (the number of times they invoke the functional composition algorithm). Based on analysis of the experimental results, we also establish some previously unknown key theoretical properties of the dominance relation directly as a function of the user preferences.

Our formalism is generic in the sense that one can use any aggregation function that appropriately represents the valuation of the composition as a function of the valuations of its constituents. In particular, we show examples of aggregation functions that compute the summation (numeric), the minimum/maximum valuation (totally ordered), or the set of *worst* valuations (partially ordered) of the constituents of a composition. Our formalism also provides flexibility in choosing the preference relation that compares sets of valuations of two compositions, so that *any* strict partial order preference relation can be used.

All our algorithms are completely independent of various aspects of the preference formalism, namely, the choice of aggregation functions, the preference relation used to compare aggregated valuations over a single attribute, and the dominance relation used to compare compositions over all attributes, except that the preference relations are strict partial orders. The theoretical and experimental results provide precise conditions under which the algorithms produce only/at least one/all of the most preferred solutions. This enables the user to choose an algorithm of his/her choice for particular problem instance, depending on the quality of solutions that is needed. In addition, our analysis also allows the user to trade off the quality of solutions produced against performance and efficiency.

## 1.4 Related Work

The closest work related to our paper is a paper by Binshtok, Brafman, Domshlak, and Shimony (2009), where preferences are expressed over collections based on the *number* of objects in the collection that satisfy a desired property (e.g., having at least two political and two sports articles in choosing articles for a newspaper publication). In contrast, we develop a formalism that considers the desirability of the collection *as a whole* based on the attributes of the objects that make up the collection, and algorithms to identify the most preferred collection(s) among those that satisfy the requirement. We further show how the problems solved using the formalism due to Binshtok et al. can also be solved in our formalism (see Section 7.3.2).

In the recent years, there has been a lot of work in the database community on the evaluation of preference queries (e.g., skyline queries) to find the most preferred subset of tuples from a result set. The problem of finding the most preferred set of tuples is analogous to finding the most preferred set of alternatives, where each alternative is a simple object, i.e., a tuple described by a set of attributes. Our problem then corresponds to finding the most preferred set of alternatives, where each alternative is in turn a *set* of tuples that *satisfy some requirement* (e.g., the set of tuples that satisfy a set of integrity constraints). Moreover, the algorithms found in the database literature mostly address totally or weakly ordered preferences over the values of attributes, while we address partially ordered preferences as well. In addition, most of them rely on the maintenance of database indexes over the attributes of the tuples because they typically cater to large scale, static data which is not typical in our setting. We however note the relevance and possible utility of techniques developed in the databases community for our problem in specific scenarios.

We refer the reader to Section 7.3 for a more detailed discussion of related work.

## 1.5 Organization

The rest of the paper is organized as follows. In Section 2, we define a compositional system, discuss the types of preferences that we will consider, and specify the problem in formal terms. In Section 3, we present our preference formalism including the dominance relation and analyze its properties. In Section 4, we present four algorithms for identifying the most preferred compositions and discuss their properties. The proofs of the results in this section are given in Appendix A. In Section 5, we discuss the complexity of our algorithms.In Section 6, we present results of experiments that we performed to compare our algorithms in terms of the quality of solutions produced, performance and efficiency. In Section 7, we summarize our contributions and discuss the related and future work in this area.

## 2. Preliminaries

We recall some basic properties and definitions concerning binary relations that we will use in the rest of the paper (see Fishburn, 1985, for a comprehensive treatment of the same).

### 2.1 Properties of Binary Relations

Let $\succ$ be a binary relation on a set $S$, i.e., $\succ \subseteq S \times S$. We say that $\succ$ is an equivalence (eq), a (strict) partial order (po), an interval order (io), a weak order (wo) or a total order (to), as defined in Table 2.

A total order is also a weak order; a weak order is also an interval order; and an interval order is also a strict partial order.

| # | Property of relation | Definition | eq | po | io | wo | to |
|---|---|---|---|---|---|---|---|
| 1. | reflexive | $\forall x \in S : x \succ x$ | ✓ | | | | |
| 2. | irreflexive | $\forall x \in S : x \nsucc x$ | | ✓ | ✓ | ✓ | ✓ |
| 3. | symmetric | $\forall x, y \in S : x \succ y \Rightarrow y \succ x$ | ✓ | | | | |
| 4. | asymmetric | $\forall x, y \in S : x \succ y \Rightarrow y \nsucc x$ | | ✓ | ✓ | ✓ | ✓ |
| 5. | transitive | $\forall x, y, z \in S : x \succ y \wedge y \succ z \Rightarrow x \succ z$ | ✓ | ✓ | ✓ | ✓ | ✓ |
| 6. | total or complete | $\forall x, y \in S : x \neq y \Rightarrow x \succ y \vee y \succ x$ | | | | | ✓ |
| 7. | negatively transitive | $\forall x, y, z \in S : x \succ y \Rightarrow x \succ z \vee z \succ y$ | | | | ✓ | ✓ |
| 8. | ferrers | $\forall x, y, z, w \in S : (x \succ y \wedge z \succ w) \Rightarrow$ $(x \succ w \vee z \succ y)$ | | | ✓ | ✓ | ✓ |

Table 2: Properties of binary relations

## 2.2 Compositional System

A compositional system consists of a repository of pre-existing components from which we are interested in assembling compositions that satisfy a pre-specified functionality. Formally, a compositional system is a tuple $\langle R, \oplus, \models \rangle$ where:

- $R = \{W_1, W_2 \ldots W_r\}$ is a set of available components,

- $\oplus$ denotes a composition operator that functionally aggregates components and encodes all the functional details of the composition. $\oplus$ is a binary operation on components $W_i, W_j$ in the repository that produces a composition $W_i \oplus W_j$.

- $\models$ is a satisfaction relation that evaluates to *true* when a composition satisfies some pre-specified functional properties.

**Definition 1** (Compositions, Feasible Compositions and Extensions). *Given a compositional system $\langle R, \oplus, \models \rangle$, and a functionality $\varphi$, a composition $\mathcal{C} = W_{i_1} \oplus W_{i_2} \oplus \ldots W_{i_n}$ is an arbitrary collection of components $W_{i_1}, W_{i_2}, \ldots, W_{i_n}$ s.t. $\forall j \in [1, n] : W_{i_j} \in R$.*

 *i. $\mathcal{C}$ is a feasible composition whenever $\mathcal{C} \models \varphi$;*

 *ii. $\mathcal{C}$ is a partial feasible composition whenever $\exists W_{j_1} \ldots W_{j_m} \in R : \mathcal{C} \oplus W_{j_1} \oplus \ldots \oplus W_{j_m}$ is a feasible composition; and*

 *iii. $\mathcal{C} \oplus W_i$ is a feasible extension of a partial feasible composition $\mathcal{C}$ whenever $\mathcal{C} \oplus W_i$ is a feasible or a partial feasible composition.*

Given a compositional system $\langle R, \oplus, \models \rangle$ and a functionality $\varphi$, an algorithm that produces a set of feasible compositions (satisfying $\varphi$) is called a *functional composition algorithm*. The most general class of functional composition algorithms we consider can be treated as *black boxes*, simply returning a set of feasible compositions satisfying $\varphi$ in a single step. Some other functional composition algorithms proceed by computing the set of feasible extensions of partial feasible compositions incrementally.

**Definition 2** (Incremental Functional Composition Algorithm). *A functional composition algorithm is said to be* incremental *if, given an initial partial feasible composition $\mathcal{C}$ and the desired functionality $\varphi$, the algorithm computes the set of feasible extensions to $\mathcal{C}$.*

An incremental functional composition algorithm can be used to compute the feasible compositions by recursively invoking the algorithm on the partial feasible compositions it produces starting with the empty composition ($\perp$), and culminating with a set of feasible compositions satisfying $\varphi$. In this sense, incremental functional composition algorithms are similar to their "black box" counterparts. However, (as we later show in Section 4.5) in contrast to their "black box" counterparts, incremental functional composition algorithms can be exploited in the search for the most preferred feasible compositions, by *interleaving* each step of the functional composition algorithm with the optimization of the valuations of non-functional attributes (with respect to the user preferences). This allows us to develop algorithms that can eliminate partial feasible compositions that will lead to less preferred feasible compositions from further consideration early in the search.

Different approaches to functional composition, (e.g., Traverso & Pistore, 2004; Lago, Pistore, & Traverso, 2002; Baier, Fritz, Bienvenu, & McIlraith, 2008; Passerone, de Alfaro, Henzinger, & Sangiovanni-Vincentelli, 2002) differ in terms of (a) the languages used to represent the desired functionality $\varphi$ and the compositions, and (b) the algorithms used to verify whether a composition $\mathcal{C}$ satisfies $\varphi$, i.e., $\mathcal{C} \models \varphi$. We have intentionally abstracted the details of how functionality $\varphi$ is represented (e.g., transition systems, logic formulas, plans, etc.) and how a composition is tested for satisfiability ($\models$) against $\varphi$, as the primary focus of our work is orthogonal to details of the specific methods used for functional composition.

## 2.3 Preferences over Non-functional Attributes

We now turn to the non-functional aspects of compositional systems. In addition to obtaining functionally feasible compositions, users are often concerned about the non-functional aspects of the compositions, e.g., the reliability of a composite Web service. In such cases, users seek the *most preferred* compositions among those that are functionally feasible, with respect to a set of non-functional attributes describing the components. In order to compute the most preferred compositions, it is necessary for the user to specify his/her preferences over a set of non-functional attributes $\mathcal{X}$.

### 2.3.1 NOTATION

In general, for any relation $\succ_P$, we use the same notation, i.e., $\succ_P$ to denote the transitive closure of the relation as well, and $\not\succ_P$ or $\neg \succ_P$ to denote its complement. The list of notations used in this paper are given in Table 3.

We focus only on *strict partial order* preference relations, i.e., relations that are both *irreflexive* and *transitive*, because transitivity is a natural property of any *rational* preference relation (von Neumann & Morgenstern, 1944; French, 1986; Mas-Colell, Whinston, & Green, 1995), and irreflexivity ensures that the preferences are strict.

With respect to any strict partial order preference relation $\succ_P$, we say that two elements $u$ and $v$ are *indifferent*, denoted $u \sim_P v$, whenever $u \not\succ_P v$ and $v \not\succ_P u$. For preference relations $\succ_i, \succ'_i, \triangleright$ and $\succ_d$, we denote the corresponding indifference relation by $\sim_i, \sim'_i, \sim_\triangleright$ and $\sim_d$ respectively. We will drop the subscripts whenever they are understood from the context.

**Proposition 1.** *For any strict partial order preference relation $\succ_P$, the corresponding indifference relation $\sim_P$ is reflexive and symmetric.*

| Notation | Meaning |
|----------|---------|
| $\mathscr{P}(S)$ | Power set of the set $S$ |
| $R = \{W_1 \cdots W_r\}$ | Set of components in the repository |
| $\oplus$ | Operation that composes components from $R$ |
| $\mathcal{C}, \mathcal{U}, \mathcal{V}, \mathcal{Z}$ | Composition or collection[4] of a set of components from $R$ |
| $\mathfrak{C}$ | A set $\{\mathcal{C}_i\}$ of compositions |
| $\mathcal{X} = \{X_1 \cdots X_m\}$ | Set of non-functional attributes |
| $\mathcal{D} = \{D_1 \cdots D_m\}$ | Set of possible valuations (domains) of attributes in $\mathcal{X}$ respectively |
| $u_i, v_i, a_i, b_i \cdots \in D_i$ | Valuations of an attribute with domain $D_i$ |
| $V_{W_i}$ | Overall valuation of the component $W_i$ with respect to all attributes $\mathcal{X}$ |
| $V_{\mathcal{C}_i}$ | Overall valuation of the composition $\mathcal{C}_i$ with respect to all attributes $\mathcal{X}$ |
| $V_{W_i}(X_j)$ | Valuation of the component $W_i$ with respect to the attribute $X_j$ |
| $V_{\mathcal{C}_i}(X_j)$ | Valuation of the composition $\mathcal{C}_i$ with respect to the attribute $X_j$ |
| $\succ_i, \succ_X$ | Intra-attribute preference over valuations of $X_i$ or $X$ respectively (user input) |
| $\rhd$ | Relative importance among attributes (user input) |
| $\Phi_i$ | Aggregation function that computes the valuation of a composition with respect to $X_i$ as a function of the valuation of its components |
| $\mathscr{F}(X_i)$ | Range of the aggregation function $\Phi_i$ for attribute $X_i$ |
| $\succ_i'$ | Derived preference relation on the aggregated valuations with respect to $X_i$ |
| $\succ_d$ | Dominance relation that compares two compositions in terms of their aggregated valuations over all attributes |
| $\Psi_\succ(S)$ | The non-dominated set of elements in $S$ with respect to $\succ$ |
| $\varphi$ | User specified functionality to be satisfied by a feasible composition |

Table 3: Notation

*Proof.* Follows from a well-known property of strict partial orders due to Fishburn (1970b). □

It is important to note that indifference with respect to a strict partial order is not necessarily transitive. For instance, $\succ_X = \{(b, c)\}$ is a strict partial order on the set $\{a, b, c\}$ with $b \sim_X a$, $a \sim_X c$ but $b \succ_X c$.

### 2.3.2 Representing Multi-Attribute Preferences

Following the representation scheme introduced by Boutilier et al. (2004) and Brafman et al. (2006), we model the user's preferences with respect to multiple attributes in two forms: (a) intra-attribute preferences with respect to each non-functional attribute in $\mathcal{X}$, and (b) relative importance over all attributes.

---

4. We will use the terms composition and collection; and component and object interchangeably.

**Definition 3** (Intra-attribute Preference). *The intra-attribute preference relation, denoted by $\succ_i$ is a strict partial order (irreflexive and transitive) over the possible valuations of an attribute $X_i \in \mathcal{X}$. $\forall u, v \in D_i : u \succ_i v$ iff $u$ is preferred to $v$ with respect to $X_i$.*

**Definition 4** (Relative Importance). *The relative importance preference relation, denoted by $\rhd$ is a strict partial order (irreflexive and transitive) over the set of all attributes $\mathcal{X}$. $\forall X_i, X_j \in \mathcal{X} : X_i \rhd X_j$ iff $X_i$ is relatively more important than $X_j$.*

Given a set $\mathcal{X}$ of attributes, the intra-attribute preference relations $\{\succ_i\}$ over their respective domains, and the relative importance preference relation $\rhd$ on $\mathcal{X}$, we address the following problems.

- Given two compositions $\mathcal{C}_j$ and $\mathcal{C}_k$, determine whether $V_{\mathcal{C}_j} \succ_d V_{\mathcal{C}_k}$ or vice versa;

- Given a compositional system $\langle R, \oplus, \models \rangle$, and an algorithm for computing a set of feasible compositions $\{\mathcal{C}_f : \mathcal{C}_f \models \varphi\}$, find the most preferred feasible compositions with respect to the above dominance relation.

## 3. Preference Formalism

Given a compositional system with a repository of components described by attributes $\mathcal{X}$ and preferences $(\{\succ_i\}, \rhd)$ over them, we are interested in reasoning about preferences over different compositions. Note that based on preferences $\{\succ_i\}$ and $\rhd$, one can make use of existing formalisms such as TCP-nets (Brafman et al., 2006) to select the most preferred components. However, the problem of comparing compositions (as opposed to comparing components) with respect to the attribute preferences is complicated by the fact that the valuation of a composition is a function of the valuations of its components. Our approach to developing the preference formalism is as follows.

First, given a composition and the valuations of its components with respect to the attributes, we obtain the *aggregated valuation* of the composition with respect to each attribute as a function of the valuations of its components. Next, we define preference relations to compare the aggregated valuations of two compositions with respect to each attribute. Finally, we build a *dominance* preference relation $\succ_d$ that *qualitatively* compares any two compositions with respect to their aggregated valuations across all attributes.

### 3.1 Aggregating Attribute Valuations across Components

In order to reason about preferences over compositions, it is necessary to obtain the valuation of a composition with respect to each attribute $X_i$ *in terms of* its components, using some *aggregation* function $\Phi_i$. There are several ways to aggregate the preference valuations attribute-wise across components in a composition. The aggregation function $\Phi_i$ defines the valuation of a composition with respect to an attribute $X_i$ as a function of the valuations of its components.

**Remark.** In the compositional systems considered here, we assume that the valuation of a composition with respect to its attributes is a function of only the valuations of its components. In other words, if $\mathcal{C} = W_1 \oplus W_2 \oplus \ldots \oplus W_n$, then $V_{\mathcal{C}}$ is a function of only $\{V_{W_1}, V_{W_2}, \ldots, V_{W_n}\}$. However, in the most general setting, the aggregation functions $\Phi_i$ need to take into account, in addition to the valuations of the components themselves, the structural or functional details of a composition encoded by $\oplus$ (e.g., the reliability of a Web service composition depends on whether the service components are composed in a series or parallel structure).

**Definition 5** (Aggregation Function). *The aggregation function on a multiset[5] of possible valuations ($D_i$) of attribute $X_i$ is*

$$\Phi_i : \mathcal{M}(D_i) \longrightarrow \mathscr{F}(X_i)$$

*where $\mathscr{F}(X_i)$ denotes the range of the aggregation function.*

Aggregation with respect to an attribute $X_i$ amounts to devising an appropriate aggregation function $\Phi_i$ that computes the valuation of a composition in terms of the valuations of its components for $X_i$. The range $\mathscr{F}(X_i)$ of $\Phi_i$ depends on the choice of aggregation function. Some examples of aggregation functions are given below.

1. *Summation.* This is applicable in cases where an attribute is real-valued and represents some kind of cost. For example, the cost of a shopping cart is the sum of the costs of the individual items it includes. In our running example, the total number of credits in a POS consisting of a set of courses is the sum of the credits of all the courses it includes. That is, if $S$ is the set of credit hours (valuations of the courses with respect to the attribute $C$) of courses in a POS, then

$$\Phi_C(S) := \{\Sigma_{s \in S} s\}$$

2. *Minimum/Maximum.* Here, the valuation of a composition with respect to an attribute is the worst, i.e., the minimum among the valuations of its components. This type of aggregation is a natural one to consider while composing embedded systems or Web services. For example, when putting together several components in an embedded system, the system is only as secure (or safe) as its least secure (or safe) component.

$$\Phi_i(S) := \{min_{s \in S} s\}$$

Analogously, one could choose as the valuation of the composition the maximum (best) among the valuations of its components. Such an aggregation function may be useful in applications such as parallel job scheduling, where the maximum response time is used to measure the quality of a schedule.

3. *Best/Worst Frontier.* In some settings, it is possible that the intra-attribute preference over the values of an attribute is a partial order (not necessarily a ranking or a total order). Hence, it may not be possible to compute the valuation of a composition as the best or worst among the valuations of its components because a unique maximum or minimum may not exist. For example, it may be useful to compute the valuation of a composition as the *minimal set* of valuations among the valuations of its components, which we call the *worst frontier*. The *worst frontier* represents the worst possible valuations of an attribute $X_i$ with respect to $\succ_i$, i.e., the minimal set[6] among the set of valuations of the components in a composition.

---

5. A multiset is a generalization of a set that allows for multiple copies of its elements.

**Definition 6** (Aggregation using Worst Frontier). *Given a set $S$ of valuations of an attribute $X_i$, the worst frontier aggregation function is defined by*

$$\forall S \subseteq D_i : \Phi_i(S) := \{v : v \in S \land \nexists u \in S : v \succ_i u\}$$

In our running example (see Section 1.1), the user would like to avoid courses not in his interest area and professors whom he is not comfortable with. That is, a program of study is considered only as good as the least interesting areas of study it covers, and the set of professors he is least comfortable with. Hence, the worst frontier aggregation function is chosen for the breadth area and instructor attributes.

***Example.*** The "worst possible" valuations of the attributes $A$ and $I$ for the program of study (composition) $P_1$ with respect to $\succ_A$ and $\succ_I$ are $\{FM, TH\}$ and $\{White, Harry\}$ respectively. Similarly, for $P_2$ the valuations of the attributes $A$ and $I$ are $\{DB, NW\}$ and $\{Jane, Tom\}$ respectively; and for $P_3$ the valuations of the attributes $A$ and $I$ are $\{CA, SE\}$ and $\{Harry, White\}$ respectively. These sets correspond to the "worst frontiers" of the respective attributes. The different areas of focus covered in the POS $P_2$ are $\{FM, AI, DB, NW, TH\}$, and the worst frontier of this set is $\Phi_A(\{FM, AI, DB, NW, TH\}) = \{DB, NW\}$ because $AI \succ_A DB, FM \succ_A DB, TH \succ_A NW$. Similarly the set of instructors in $P_2$ are $\{Tom, Gopal, Bob, Jane\}$, and hence we have $\Phi_I(\{Tom, Gopal, Bob, Jane\}) = \{Jane, Tom\}$ because $Bob \succ_I Jane$ and $Gopal \succ_I Tom$. For attribute $C$, the aggregation function evaluates the sum of credits of the constituent courses in a POS. Therefore, for $P_2$ we have $\Phi_C(\{4, 3, 4, 2, 3, 3\}) = 4 + 3 + 4 + 2 + 3 + 3 = 19$. $\diamond$

We note that other choices of the aggregation function can be accommodated in our framework (such as average or a combination of best and worst frontier sets), and that the above is only a representative list of choices.

**Proposition 2** (Indifference of Frontier Elements). *Consider an attribute $X_i$, whose valuations are aggregated using the best or worst frontier aggregation function. Let $A \in \mathscr{F}(X_i)$. Then $u \sim_i v$ for all $u, v \in A$.*

*Proof.* Follows from Definition 6 (or the analogous definition of a best frontier) and a well-known result due to the work of Fishburn (1985). $\blacksquare$

**Definition 7** (Valuation of a Composition for Attributes Aggregated using Best/Worst Frontier). *Consider an attribute $X_i$, whose valuations are aggregated using the best or worst frontier aggregation function. The valuation of a component $W$ with respect to an attribute $X_i$ is denoted as $V_W(X_i) \in D_i$. The valuation of a composition of two components $W_1$ and $W_2$ with respect to an attribute $X_i$, each with valuation $V_{W_1}(X_i)$ and $V_{W_2}(X_i)$ respectively, is given by*

---

6. Note that if $\succ_i$ is a total order, then worst frontier represents the minimum or lowest element in the set with respect to the total order.

$$V_{W_1 \oplus W_2}(X_i) := \Phi_i(V_{W_1}(X_i) \cup V_{W_2}(X_i))$$

**Example.** Consider $P_2 = CS501 \oplus CS502 \oplus CS505 \oplus CS506 \oplus CS509 \oplus CS510$ in our running example (see Section 1.1).

$$
\begin{aligned}
V_{P_2}(I) &= \Phi_I(V_{CS501}(I) \cup V_{CS502}(I) \cup V_{CS505}(I) \cup V_{CS506}(I) \cup V_{CS509}(I) \cup V_{CS510}(I)) \\
&= \Phi_I(\{Tom\} \cup \{Gopal\} \cup \{Bob\} \cup \{Bob\} \cup \{Jane\} \cup \{Tom\}) \\
&= \Phi_I(\{Tom, Gopal, Bob, Jane\}) \\
&= \{Tom, Jane\}
\end{aligned}
$$

$\diamond$

It must be noted that $V_{W_1 \oplus W_2}(X_i) = V_{W_2 \oplus W_1}(X_i)$ according to the above definition, because the valuations of compositions are *subsets* of the union of individual component valuations.

## 3.2 Comparing Aggregated Valuations

Having obtained an aggregated valuation with respect to each attribute, we next proceed to discuss how to compare aggregated valuations attribute-wise. We denote the preference relation used to compare the aggregated valuations for an attribute $X_i$ by $\succ_i'$. In the simple case when an aggregation function $\Phi_i$ with respect to an attribute $X_i$ returns a value in $D_i$ ($\mathscr{F}(X_i) = D_i$), the intra-attribute preference $\succ_i$ can be (re)used to compare aggregated valuations, i.e., $\succ_i' = \succ_i$. Other choices of $\succ_i'$ can be considered as long as $\succ_i'$ is a partial order. In order to obtain a strict preference relation, we require irreflexivity, and to obtain a rational preference relation, we require transitivity[7].

For worst frontier-based aggregation (Definition 6), we present a preference relation that uses the following idea: Given two compositions with different aggregated valuations (worst frontiers) $A, B$ with respect to an attribute $X_i$, we say that $A$ is preferred to $B$ if for *every* valuation of $X_i$ in $B$, there is *some* valuation in $A$ that is strictly preferred.

**Definition 8** (Preference over Worst Frontiers). *Let $A, B \in \mathscr{F}(X_i)$ be two worst frontiers with respect to attribute $X_i$. We say that valuation $A$ is preferred to $B$ with respect to $X_i$, denoted by $A \succ_i' B$, if for each element in $B$, there exists an element in $A$ that is more preferred.*

$$\forall A, B \in \mathscr{F}(X_i) : A \succ_i' B \Leftrightarrow \forall b \in B, \exists a \in A : a \succ_i b$$

**Example.** In our running example (see Section 1.1), we have $\{FM, TH\} \succ_A' \{DB, NW\}$ because $FM \succ_A DB$ and $TH \succ_A NW$. $\diamond$

Given a preference relation over a set of elements, there are several ways of obtaining a preference relation over subsets of elements from the set (see Barbera, Bossert, & Pattanaik, 2004, for a survey on preferences over sets). Definition 8 is one simple way to achieve this. In some settings, in contrast to Definition 8, it might be useful to compare only elements in the two sets that are not common. In such settings, a suitable irreflexive and transitive preference relation can be used, such as the asymmetric part of preference relations developed by Brewka et al. (2010) and Bouveret et al. (2009). In the absence of transitivity, the transitive closure of the relation may be used to compare sets of elements, as done by Brewka et al.

---

7. Any preference relation, including the one that compares only the uncommon elements of two sets can be used, provided it is irreflexive and transitive.

We now discuss some properties of the specific relation $\succ_i'$ as introduced in Definition 8.

**Proposition 3** (Irreflexivity of $\succ_i'$). $A \in \mathscr{F}(X_i) \Rightarrow A \not\succ_i' A$.

*Proof.* $\forall a, b \in A, a \sim_i b$ (follows from *Proposition 2*) □

**Proposition 4** (Transitivity of $\succ_i'$). *If $A, B, C \in \mathscr{F}(X_i)$, then $A \succ_i' B \wedge B \succ_i' C \Rightarrow A \succ_i' C$.*

*Proof.* Immediate from Definition 8. □

**Definition 9.** *Let $A, B \in \mathscr{F}(X_i)$. We say that valuation $A$ is* at least as preferred *as $B$ with respect to $X_i$, denoted $\succeq_i'$ iff*

$$A \succeq_i' B \Leftrightarrow A = B \vee A \succ_i' B$$

**Proposition 5.** $\succeq_i'$ *is reflexive and transitive.*

*Proof.* Follows from the facts that $=$ is reflexive and transitive, and $\succ_i'$ is irreflexive and transitive. □

**Definition 10** (Complete Valuation). *The complete valuation or outcome or assignment of a composition $\mathcal{C}$ is defined as a tuple $V_{\mathcal{C}} := \langle V_{\mathcal{C}}(X_1), \dots V_{\mathcal{C}}(X_m) \rangle$, where $V_{\mathcal{C}}(X_i) \in \mathscr{F}(X_i)$. The set of all possible valuations or outcomes is denoted as $\prod_{i=1}^{m} \mathscr{F}(X_i)$.*

***Example.*** In case of our example in Section 1.1:

$$
\begin{aligned}
V_{P_1} &= \langle \Phi_A(\{FM, AI, TH\}), \Phi_I(\{Tom, Gopal, Harry, White, Jane\}), \Phi_C(\{4, 3, 2, 3, 3, 3\}) \rangle \\
&= \langle \{FM, TH\}, \{White, Harry\}, \{18\} \rangle \\
V_{P_2} &= \langle \Phi_A(\{FM, AI, DB, NW, TH\}), \Phi_I(\{Tom, Gopal, Bob, Jane\}), \Phi_C(\{4, 3, 4, 2, 3, 3\}) \rangle \\
&= \langle \{DB, NW\}, \{Tom, Jane\}, \{19\} \rangle \\
V_{P_3} &= \langle \Phi_A(\{FM, AI, CA, SE, TH\}), \Phi_I(\{Harry, White, Tom, Jane\}), \Phi_C(\{2, 3, 3, 2, 3, 3\}) \rangle \\
&= \langle \{CA, SE\}, \{Harry, White\}, \{16\} \rangle
\end{aligned}
$$

$\diamond$

## 3.3 Dominance: Preference over Compositions

In the previous sections, we have discussed how to evaluate and compare a composition with respect to the attributes as a function of its components. In order to identify preferred compositions, we need to compare compositions with respect to their aggregated valuations over all attributes, based on the originally specified intra-attribute and relative importance preferences. We present a specific dominance relation for performing such a comparison.

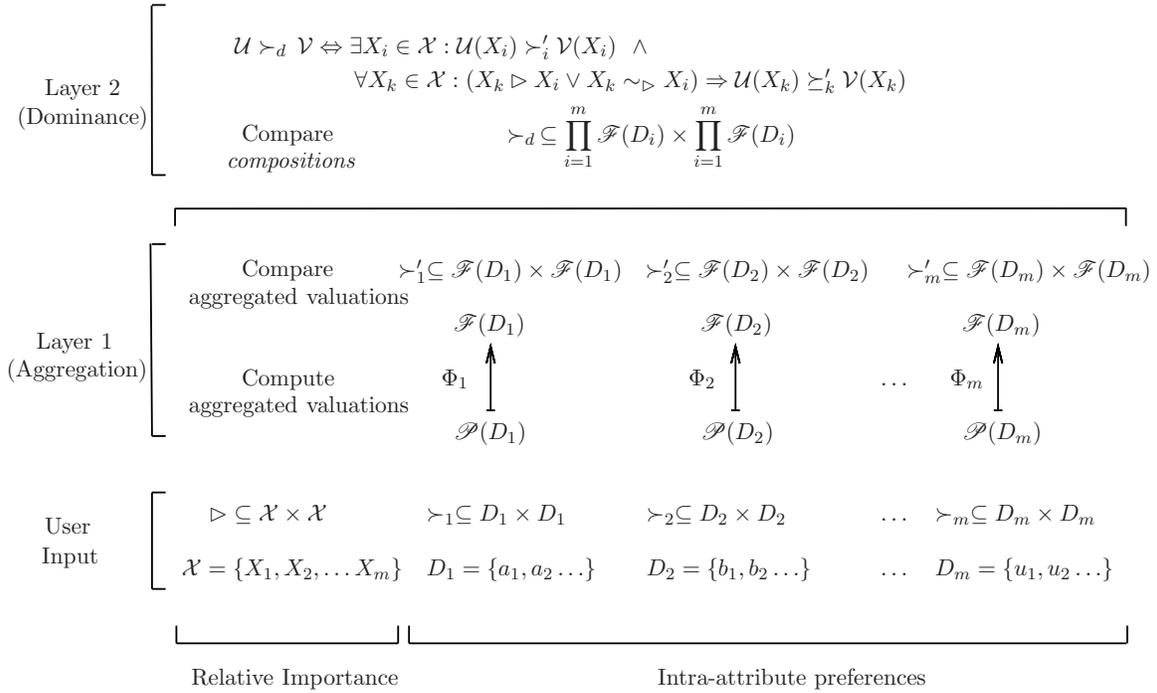

Figure 2: Dominance: Preference over compositions

**Definition 11** (Dominance). *Dominance* $\succ_d$ *is a binary relation defined as follows: for all* $\mathcal{U}^8, \mathcal{V} \in \prod_{i=1}^{m} \mathscr{F}(X_i)$

$$\mathcal{U} \succ_d \mathcal{V} \Leftrightarrow \quad \exists X_i \; : \; \mathcal{U}(X_i) \succ_i' \mathcal{V}(X_i) \; \wedge$$
$$\forall X_k \; : \; (X_k \rhd X_i \vee X_k \sim_\rhd X_i) \; \Rightarrow \; \mathcal{U}(X_k) \succeq_k' \mathcal{V}(X_k)$$

In Definition 11, we call the attribute $X_i$ as the "*witness*" of the relation. The dominance relation $\succ_d$ is derived from and respects both the intra-attribute preferences ($\succ_i$) as well as the relative importance preferences ($\rhd$) asserted by the user. Figure 2 graphically illustrates how dominance is derived from user-specified preferences. First, to start with we have user specified preferences, namely intra-attribute ($\succ_i$) and relative importance ($\rhd$) preferences. Next, from $\succ_i$ preferences, the valuations of compositions with respect to attributes are computed using the aggregation function ($\Phi_i$). Then the intra-attribute preference relation to compare the aggregated valuations ($\succ_i'$) is derived from $\succ_i$. Finally, the global dominance ($\succ_d$) is defined in terms of $\succ_i'$ and $\rhd$.

The definition of dominance states that a composition $\mathcal{U}$ dominates $\mathcal{V}$ iff we can find a witness attribute $X_i$ such that with respect to the intra-attribute preference $\succ_i$, the valuation of $\mathcal{U}$ dominates $\mathcal{V}$ in terms of $\succ_i'$, and for all attributes $X_k$ which the user considers more important than ($\rhd$) or indifferent to ($\sim_\rhd$) $X_i$, the valuation of $X_k$ in $\mathcal{U}$ is at least as preferred ($\succeq_i'$) as the valuation of $X_k$ in $\mathcal{V}$.

---

8. To avoid excessively cluttering the notation, for a given composition $\mathcal{C}$, we will slightly abuse notation by using $\mathcal{C}$ interchangeably with $V_\mathcal{C}$.

***Example.*** In our running example (see Section 1.1), we have $V_{P_2} \succ_d V_{P_1}$ with $I$ as witness and $V_{P_1} \succ_d V_{P_3}$ with $A$ as witness. If $I \triangleright A$, $I \triangleright C$ but $A \sim_{\triangleright} C$ then $V_{P_2} \succ_d V_{P_1}$ and $V_{P_2} \succ_d V_{P_3}$ with $I$ as witness, but $V_{P_1} \not\succ_d V_{P_3}$ and $V_{P_3} \not\succ_d V_{P_1}$. This is because $P_1$ is preferred to $P_3$ with respect to $A$ ($\{FM, TH\} \succ'_A \{CA, SE\}$); but $P_3$ is preferred to $P_1$ with respect to $C$ ($\{16\} \succ'_C \{18\}$), and neither $A$ nor $C$ is relatively more important than the other. $\diamond$

### 3.4 Properties of $\succ_d$

We now proceed to analyze some properties of $\succ_d$ with respect to the worst-frontier aggregation function. First, we show that a partial feasible composition is not dominated with respect to $\succ_d$ by any of its extensions. This property will be useful in establishing the soundness of algorithms that compute the most preferred compositions (see Section 4). Next, we observe that $\succ_d$ is irreflexive (follows from the irreflexivity of $\succ_i$), and proceed to identify the conditions under which $\succ_d$ is transitive. We focus on transitive preferences because many studies have considered transitivity to be a key property of preference relations (von Neumann & Morgenstern, 1944; French, 1986; Mas-Colell et al., 1995)[9].

**Proposition 6.** *Whenever preferences are aggregated using the worst-frontier based aggregation function, for any partial feasible composition $\mathcal{C}$, there is no feasible extension $\mathcal{C} \oplus W$ that dominates it, i.e., $V_{\mathcal{C} \oplus W} \not\succ_d V_{\mathcal{C}}$.*

*Proof.* The proof proceeds by showing that with respect to each attribute $X_i$, $V_{\mathcal{C} \oplus W}(X_i) \not\succ'_i V_{\mathcal{C}}(X_i)$, thereby ruling out the existence of a witness for $V_{\mathcal{C} \oplus W} \succ_d V_{\mathcal{C}}$. Suppose that by contradiction, $\mathcal{C} \oplus W$ is a feasible extension of $\mathcal{C}$ such that $V_{\mathcal{C} \oplus W} \succ_d V_{\mathcal{C}}$. By Definition 11, $V_{\mathcal{C} \oplus W} \succ_d V_{\mathcal{C}}$ requires the existence of a witness attribute $X_i \in \mathcal{X}$ such that $V_{\mathcal{C} \oplus W}(X_i) \succ'_i V_{\mathcal{C}}(X_i)$, i.e.,

$$\forall b \in V_{\mathcal{C}}(X_i) \ \exists a \in V_{\mathcal{C} \oplus W}(X_i) : a \succ_i b \tag{1}$$

By Definition 7, we have $V_{\mathcal{C} \oplus W}(X_i) = \Phi_i(V_{\mathcal{C}}(X_i) \cup V_W(X_i))$. However, by Definition 6 $a \in \Phi_i(V_{\mathcal{C}}(X_i) \cup V_W(X_i)) \Rightarrow \not\exists b \in V_{\mathcal{C}}(X_i) \cup V_W(X_i) : a \succ_i b$, which contradicts Equation (1). This rules out the existence of a witness for $V_{\mathcal{C} \oplus W} \succ_d V_{\mathcal{C}}$. Hence, $V_{\mathcal{C} \oplus W} \not\succ_d V_{\mathcal{C}}$. $\square$

We next proceed to show that $\succ_d$ is not necessarily transitive when intra-attribute and relative importance preference relations are both arbitrary strict partial orders.

**Proposition 7.** *When intra-attribute preferences $\succ_i$ as well as relative importance among attributes $\triangleright$ are arbitrary partial orders, $\mathcal{U} \succ_d \mathcal{V} \wedge \mathcal{V} \succ_d \mathcal{Z} \not\Rightarrow \mathcal{U} \succ_d \mathcal{Z}$*

*Proof.* We show a counter example of a compositional system with partially ordered $\{\succ_i\}, \triangleright$ and compositions $\mathcal{U}, \mathcal{V}, \mathcal{Z}$ such that $\mathcal{U} \succ_d \mathcal{V}$, $\mathcal{V} \succ_d \mathcal{Z}$ but $\mathcal{U} \not\succ_d \mathcal{Z}$.

Consider a system with a set of attributes $\mathcal{X} = \{X_1, X_2, X_3, X_4\}$, each with domains $D_1 = \{a_1, b_1\}, \ldots D_4 = \{a_4, b_4\}$. Let the relative importance relation $\triangleright$ on $\mathcal{X}$ and the intra-attribute preferences $\succ_1 \ldots \succ_4$ be given by $\triangleright = \{(X_1, X_3), (X_2, X_4)\}$ and $\succ_i = \{(a_i, b_i)\}, i = 1, 2, 3, 4$ respectively (Figure 3). The valuations of $\mathcal{U}, \mathcal{V}, \mathcal{Z}$ with respect to the attributes $\mathcal{X}$ are given in Table 4.

---

9. While some studies of human decision making have argued that human preferences are not necessarily transitive (Tversky, 1969), others have offered evidence to the contrary (Regenwetter, Dana, & Davis-Stober, 2011).

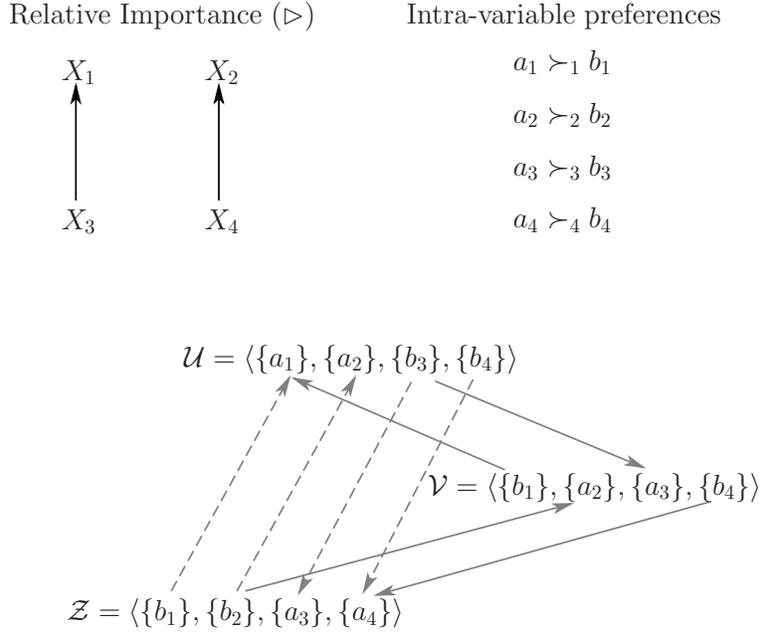

Figure 3: Counter example

| Comp. $(\mathcal{C})$ | $V_{\mathcal{C}}(X_1)$ | $V_{\mathcal{C}}(X_2)$ | $V_{\mathcal{C}}(X_3)$ | $V_{\mathcal{C}}(X_4)$ |
| :---: | :---: | :---: | :---: | :---: |
| $\mathcal{U}$ | $a_1$ | $a_2$ | $b_3$ | $b_4$ |
| $\mathcal{V}$ | $b_1$ | $a_2$ | $a_3$ | $b_4$ |
| $\mathcal{Z}$ | $b_1$ | $b_2$ | $a_3$ | $a_4$ |

Table 4: Valuations of $\mathcal{U}, \mathcal{V}, \mathcal{Z}$

Clearly $\mathcal{U} \succ_d \mathcal{V}$ with $X_1$ as the witness, and $\mathcal{V} \succ_d \mathcal{Z}$ with $X_2$ as the witness. In addition, note that:

$$\mathcal{Z}(X_3) \succ'_3 \mathcal{U}(X_3) \tag{2}$$

$$\mathcal{Z}(X_4) \succ'_4 \mathcal{U}(X_4) \tag{3}$$

However, we observe that $\mathcal{U} \not\succ_d \mathcal{Z}$:

a. $X_1$ is not a witness due to $X_4 \sim_\rhd X_1$ and Equation (3).

b. $X_2$ is not a witness due to $X_3 \sim_\rhd X_2$ and Equation (2).

c. $X_3$ is not a witness due to Equation (2).

d. $X_4$ is not a witness due to Equation (3). $\qquad\square$

The above proposition shows that the dominance relation $\succ_d$ is not transitive when $\succ_i$ and $\rhd$ are arbitrary partial orders, when considering worst-frontier based aggregation. Because transitivity of preference is a necessary condition for rational choice (von Neumann

& Morgenstern, 1944; French, 1986; Mas-Colell et al., 1995), we proceed to investigate the possibility of obtaining such a dominance relation by restricting $\triangleright$. We later prove that such a restriction is necessary and sufficient for the transitivity of $\succ_d$.

**Definition 12** (Relative Importance as an Interval Order). *A relative importance relation* $\triangleright$ *is a binary relation which is reflexive and satisfies the following axiom.*

$$\forall X_i, X_j, X_k, X_l \in \mathcal{X} : (X_i \triangleright X_j \wedge X_k \triangleright X_l) \Rightarrow (X_i \triangleright X_l \vee X_k \triangleright X_j) \tag{4}$$

*We say that $X_i$ is relatively more important than $X_j$ if $X_i \triangleright X_j$.*

**Proposition 8** (Transitivity of $\triangleright$ see Fishburn, 1985). $\triangleright$ *is transitive.* $\qquad\blacksquare$

**Remarks.**

1. Definition 12 imposes an additional restriction on the structure of the relative importance relation $\triangleright$, over a strict partial order. A strict partial order is just irreflexive and transitive; however, the relative importance relation in Definition 12 should in addition satisfy Equation (4), thereby yielding an *interval order* (Fishburn, 1985).

2. The indifference relation with respect to $\triangleright$, namely $\sim_\triangleright$ is *not* transitive. For example, if there are three attributes $\mathcal{X} = \{X_1, X_2, X_3\}$, and $\triangleright = \{(X_1, X_2)\}$. $\triangleright$ satisfies the condition for an *interval order*, and we have $X_1 \sim_\triangleright X_3$ and $X_3 \sim_\triangleright X_2$, but $X_1 \not\sim_\triangleright X_2$ because $X_1 \triangleright X_2$.

Propositions 9-12 establish the properties of the dominance relation $\succ_d$ in the case where the relative importance relation $\triangleright$ is an interval order. In particular, we prove that $\succ_d$ is irreflexive (Proposition 9) and transitive (Proposition 12), making $\succ_d$ a *strict partial order* (Theorem 1).

**Proposition 9** (Irreflexivity of $\succ_d$). $\mathcal{U} \in \prod_{i=1}^{m} \mathscr{F}(X_i) \Rightarrow \mathcal{U} \not\succ_d \mathcal{U}.$

*Proof.* Suppose that $\mathcal{U} \succ_d \mathcal{U}$ by contradiction. Then $\exists X_i, \ s.t. \ \mathcal{U}(X_i) \succ_i' \mathcal{U}(X_i)$ by definition. But this contradicts Proposition 3. $\qquad\blacksquare$

The above proposition ensures that the dominance relation $\succ_d$ is strict over compositions. In other words, no composition is preferred over itself. Next, we proceed to establish the other important property of rational preference relations: transitivity of $\succ_d$. We make use of two intermediate propositions 10 and 11 that are needed for the task.

In Proposition 10, we prove that if an attribute $X_i$ is relatively more important than $X_j$, then $X_i$ is not more important than a third attribute $X_k$ implies that $X_j$ is also not more important than $X_k$. This will help us prove the transitivity of the dominance relation. Figure 4 illustrates the cases that arise.

**Proposition 10.** $\forall X_i, X_j, X_k :$
$X_i \triangleright X_j \Rightarrow \Big( (X_k \triangleright X_i \vee X_k \sim_\triangleright X_i) \Rightarrow (X_k \triangleright X_j \vee X_k \sim_\triangleright X_j) \Big)$

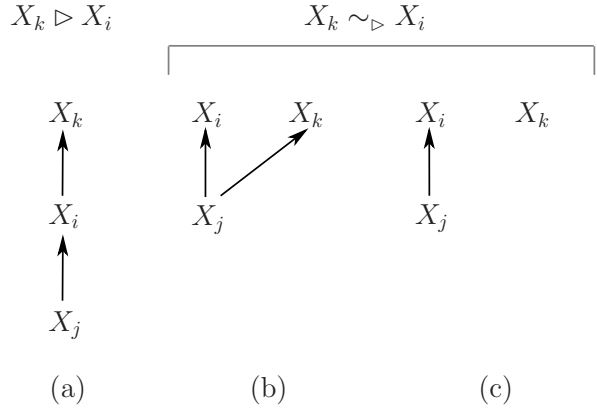

Figure 4: $X_i \rhd X_j \wedge (X_k \rhd X_i \vee X_k \sim_\rhd X_i)$

The proof follows from the fact that $\rhd$ is a partial order.

*Proof.*

1. $X_i \rhd X_j$   (*Hyp.*)

2. $X_k \rhd X_i \vee X_k \sim_\rhd X_i$   (*Hyp.*) Show $X_k \rhd X_j \vee X_k \sim_\rhd X_j$

    2.1. $X_k \rhd X_i \Rightarrow X_k \rhd X_j$   By transitivity of $\rhd$ and (1.); see Figure 4(a)

    2.2. $X_k \sim_\rhd X_i \Rightarrow X_k \rhd X_j \vee X_k \sim_\rhd X_j$

        i. $X_k \sim_\rhd X_i$   (*Hyp.*)

        ii. $(X_k \rhd X_j) \vee (X_j \rhd X_k) \vee (X_k \sim_\rhd X_j)$   Always; see Figure 4(b,c)

        iii. $X_j \rhd X_k \Rightarrow X_i \rhd X_k$   (1.) Contradiction!

        iv. $X_k \rhd X_j \vee X_k \sim_\rhd X_j$   (2.2.*ii.,iii.*)

3. $X_i \rhd X_j \Rightarrow \Big((X_k \rhd X_i \vee X_k \sim_\rhd X_i) \Rightarrow (X_k \rhd X_j \vee X_k \sim_\rhd X_j)\Big)$   (1., 2.1, 2.2)   □

Proposition 11 states that if attributes $X_i, X_j$ are such that $X_i \sim_\rhd X_j$ then at least one of them, $X_u$ is such that with respect to the other, $X_v$, there is no attribute $X_k$ that is less important while at the same time $X_k \sim_\rhd X_u$. This result is needed to establish the transitivity of the dominance relation.

**Proposition 11.** $\forall X_i, X_j, u \neq v, X_i \sim_\rhd X_j \Rightarrow \exists X_u, X_v \in \{X_i, X_j\}, \nexists X_k : (X_u \sim_\rhd X_k \wedge X_v \rhd X_k)$

The proof makes use of the fact that relative importance is an interval order relation.

*Proof.* Let $X_i \sim_\rhd X_j$, and $X_i'$ and $X_j'$ be attributes that are less important than $X_i$ and $X_j$ respectively (if any). Figure 5 illustrates all the cases. Figure 5(a, b, c, d, e) illustrates the cases when *at most* one of $X_i'$ and $X_j'$ exists, and in each case the claim holds trivially. For example, in the cases of Figure 5(a, b, c), both $X_u = X_i; X_v = X_j$ and $X_u = X_j; X_v = X_i$ satisfy the implication, and in the cases of Figure 5(d, e), the corresponding satisfactory assignments to $X_u$ and $X_v$ are shown in the figure. The case of Figure 5(f) never arises because $\rhd$ is an interval order (see Definition 12). Hence, the proposition holds in all cases.   □

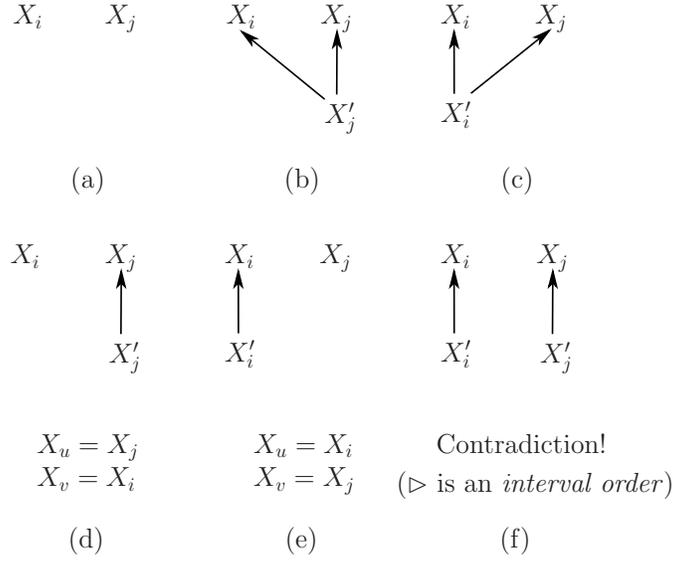

Figure 5: $X_i \sim_\rhd X_j$

The above proposition reflects the interval order property of the $\rhd$ relation, and it complements the result of Proposition 7, where $\succ_d$ was shown to be intransitive when $\rhd$ is not an interval order. In fact, if relative importance was defined as a strict partial order instead, the above proof does not hold. Given that $\mathcal{U} \succ_d \mathcal{V}$ with witness $X_i$ and $\mathcal{V} \succ_d \mathcal{Z}$ with witness $X_j$, the above proposition guarantees that one among $X_i$ and $X_j$ can be chosen as a potential witness for $\mathcal{U} \succ_d \mathcal{Z}$ so that the conditions demonstrated in the counter example of Proposition 7 are avoided. Using the propositions 10 and 11, we are now in a position to prove transitivity of $\succ_d$ in Proposition 12.

**Proposition 12** (Transitivity of $\succ_d$). $\forall\, \mathcal{U}, \mathcal{V}, \mathcal{Z} \in \prod_{i=1}^{m} \mathscr{F}(X_i)$,

$\mathcal{U} \succ_d \mathcal{V} \wedge \mathcal{V} \succ_d \mathcal{Z} \Rightarrow \mathcal{U} \succ_d \mathcal{Z}$.

The proof proceeds by considering all possible relationships between $X_i, X_j$, the respective attributes that are *witnesses* of the dominance of $\mathcal{U}$ over $\mathcal{V}$ and $\mathcal{V}$ over $\mathcal{Z}$. Lines $5, 6, 7$ in the proof establish the dominance of $\mathcal{U}$ over $\mathcal{Z}$ in the cases $X_i \rhd X_j$, $X_j \rhd X_i$ and $X_i \sim_\rhd X_j$ respectively. In the first two cases, the more important attribute among $X_i$ and $X_j$ is shown to be the witness for $\mathcal{U} \succ_d \mathcal{Z}$ with the help of Proposition 10; and in the last case we make use of Proposition 11 to show that at least one of $X_i, X_j$ is a witness for $\mathcal{U} \succ_d \mathcal{Z}$.

*Proof.*

1. $\mathcal{U} \succ_d \mathcal{V}$   (*Hyp.*)

2. $\mathcal{V} \succ_d \mathcal{Z}$   (*Hyp.*)

3. $\exists X_i \;:\; \mathcal{U}(X_i) \succ_i' \mathcal{V}(X_i)$   (1.)

4. $\exists X_j \;:\; \mathcal{V}(X_j) \succ_j' \mathcal{Z}(X_j)$   (2.)
   Three cases arise: $X_i \rhd X_j (5.)$, $X_j \rhd X_i (6.)$ and $X_i \sim_\rhd X_j (7.)$.

5. $X_i \rhd X_j \Rightarrow \mathcal{U} \succ_d \mathcal{Z}$

   5.1. $X_i \rhd X_j$   (*Hyp.*)

   5.2. $\mathcal{V}(X_i) \succeq'_i \mathcal{Z}(X_i)$   (2., 5.1.)

   5.3. $\mathcal{U}(X_i) \succ'_i \mathcal{Z}(X_i)$   (3., 5.2.)

   5.4. $\forall X_k : (X_k \rhd X_i \vee X_k \sim_\rhd X_i) \Rightarrow \mathcal{U}(X_k) \succeq'_k \mathcal{Z}(X_k)$

      i. Let $X_k \rhd X_i \vee X_k \sim_\rhd X_i$   (*Hyp.*)

      ii. $\mathcal{U}(X_k) \succeq'_k \mathcal{V}(X_k)$   (1., 5.4.*i.*)

      iii. $X_k \rhd X_j \vee X_k \sim_\rhd X_j$   (5.4.*i.*, *Proposition* 10)

      iv. $\mathcal{V}(X_k) \succeq'_k \mathcal{Z}(X_k)$   (2., 5.4.*iii.*)

      v. $\mathcal{U}(X_k) \succeq'_k \mathcal{Z}(X_k)$   (5.4.*ii.*, 5.4.*iv.*)

   5.5. $X_i \rhd X_j \Rightarrow \mathcal{U} \succ_d \mathcal{Z}$   (5.1., 5.3., 5.4.)

6. $X_j \rhd X_i \Rightarrow \mathcal{U} \succ_d \mathcal{Z}$

   6.1. This is true by symmetry of $X_i, X_j$ in the proof of (5.); in this case, it can easily be shown that $\mathcal{U}(X_j) \succ'_i \mathcal{Z}(X_j)$ and $\forall X_k : (X_k \rhd X_j \vee X_k \sim_\rhd X_j) \Rightarrow \mathcal{U}(X_k) \succeq'_k \mathcal{Z}(X_k)$.

7. $X_i \sim_\rhd X_j \Rightarrow \mathcal{U} \succ_d \mathcal{Z}$

   7.1. $X_i \sim_\rhd X_j$   (*Hyp.*)

   7.2. $\exists X_u, X_v \in \{X_i, X_j\} : X_u \neq X_v \wedge \nexists X_k : (X_u \sim_\rhd X_k \wedge X_v \rhd X_k)$   (7.1., *Proposition* 11)

   7.3. Without loss of generality, suppose that $X_u = X_i, X_v = X_j$   (*Hyp.*).

   7.4. $\mathcal{V}(X_i) \succeq'_i \mathcal{Z}(X_i)$   (2., 7.1.)

   7.5. $\mathcal{U}(X_i) \succ'_i \mathcal{Z}(X_i)$   (3., 7.4.)

   7.6. $\forall X_k : X_k \rhd X_i \Rightarrow \mathcal{U}(X_k) \succeq'_k \mathcal{Z}(X_k)$.

      i. $X_k \rhd X_i$   (*Hyp.*)

      ii. $\mathcal{U}(X_k) \succeq'_k \mathcal{V}(X_k)$   (1., 7.6.*i.*)

      iii. $X_k \rhd X_j \vee X_k \sim_\rhd X_j$   Because $X_j \rhd X_k$ Contradicts (7.1., 7.6.*i.*)!

      iv. $\mathcal{V}(X_k) \succeq'_k \mathcal{Z}(X_k)$   (2., 7.6.*iii.*)

      v. $\mathcal{U}(X_k) \succeq'_k \mathcal{Z}(X_k)$   (7.6.*ii.*, 7.6.*iv.*)

   7.7. $\forall X_k : X_k \sim_\rhd X_i \Rightarrow \mathcal{U}(X_k) \succeq'_k \mathcal{Z}(X_k)$

      i. $X_k \sim_\rhd X_i$   (*Hyp.*)

      ii. $\mathcal{U}(X_k) \succeq'_k \mathcal{V}(X_k)$   (1., 7.7.*i.*)

      iii. $X_k \rhd X_j \vee X_k \sim_\rhd X_j$   Because $X_j \rhd X_k$ Contradicts (7.2., 7.3.)!

      iv. $\mathcal{V}(X_k) \succeq'_k \mathcal{Z}(X_k)$   (2., 7.7.*iii.*)

      v. $\mathcal{U}(X_k) \succeq'_k \mathcal{Z}(X_k)$   (7.7.*ii.*, 7.7.*iv.*)

   7.8. $\forall X_k : X_k \rhd X_i \vee X_k \sim_\rhd X_i \Rightarrow \mathcal{U}(X_k) \succeq'_k \mathcal{Z}(X_k)$   (7.6., 7.7.)

   7.9. $X_i \sim_\rhd X_j \Rightarrow \mathcal{U} \succ_d \mathcal{Z}$   (7.5., 7.8.)

8. $(X_i \rhd X_j \vee X_j \rhd X_i \vee X_i \sim_\rhd X_j) \Rightarrow \mathcal{U} \succ_d \mathcal{Z}$   (5., 6., 7.)

9. $\mathcal{U} \succ_d \mathcal{V} \wedge \mathcal{V} \succ_d \mathcal{Z} \Rightarrow \mathcal{U} \succ_d \mathcal{Z}$   (1., 2., 8.)      $\square$

**Theorem 1.** *If the intra-attribute preferences $\succ_i$ are arbitrary strict partial orders and relative importance $\rhd$ is an interval order, then $\succ_d$ is a strict partial order.*

*Proof.* Follows immediately from Propositions 9 and 12.      $\square$

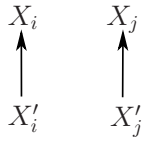

Figure 6: A $2 \oplus 2$ substructure, not an Interval Order

## 3.5 Role of Interval Order Restriction on $\rhd$ in the Transitivity of $\succ_d$

Theorem 1 establishes that given partially ordered intra-attribute preferences $\succ_i$, if the relative importance relation ($\rhd$) is an interval order (Definition 12), then $\succ_d$ is transitive. In addition, we have also seen a counter example in Proposition 7, which shows that the transitivity of $\succ_d$ does not necessarily hold when $\rhd$ is an arbitrary partial order.

Is there a condition *weaker* than the interval order restriction that still makes $\succ_d$ transitive when retain intra-attribute preferences as arbitrary partial orders and dominance as in Definition 11? The answer turns out to be 'no', which we prove next.

Before we proceed to prove the necessity of an interval ordered relative importance relation $\rhd$ for a transitive dominance relation $\succ_d$, we will examine interval orders more closely. Recall from Definition 12 that every interval order $\rhd$ on $\mathcal{X}$ is a partial order, and it additionally satisfies Ferrer's axiom for all $X_1, X_2, X_3, X_4 \in \mathcal{X}$:

$$(X_1 \rhd X_2 \wedge X_3 \rhd X_4) \Rightarrow (X_1 \rhd X_4 \vee X_3 \rhd X_2)$$

We borrow a characterization of the above axiom by Fishburn (1970a, 1985) that the relation $\rhd$ is an interval order if and only if $2 \oplus 2 \nsubseteq \rhd$, where $2 \oplus 2$ is a relational structure shown in Figure 6. In other words, a partial order is an interval order if and only if it has *no restriction of itself* that is isomorphic to the partial order structure shown in Figure 6.

**Theorem 2** (Necessity of Interval Order)**.** *For partially ordered intra-attribute preferences and dominance relation in Definition 11, $\succ_d$ is transitive only if relative importance $\rhd$ is an interval order.*

*Proof.* Assume that $\rhd$ is not an interval order. This is true if and only if $2 \oplus 2 \subseteq \rhd$. However, we showed in Proposition 7 that in this case, $\succ_d$ is not transitive using a counter example (see Figure 3). Hence, $\succ_d$ is transitive only if relative importance $\rhd$ is an interval order. □

## 3.6 Additional Properties of $\succ_d$ with Respect to the Properties of $\{\succ_i\}$ and $\rhd$

We now present some additional properties[10] of $\succ_d$ that hold when certain restrictions are imposed on the intra-attribute and relative importance preference relations.

**Proposition 13.** *If $\rhd$ is a total order and $X_i$ is the most important attribute in $\mathcal{X}$ with respect to $\rhd$, then $\succ'_i \subseteq \succ_d$.*

---

10. The results in this section essentially prove conjectures that arose out of analysis of the results of our experiments (see Section 6).

*Proof.* Let $X_i$ be the (unique) most important attribute in $\mathcal{X}$. Suppose that $\mathcal{U}(X_i) \succ'_i \mathcal{V}(X_i)$, thereby making $X_i$ a potential witness for $\mathcal{U} \succ_d \mathcal{V}$. Since $X_i$ is the most important attribute, $\forall X_k \in \mathcal{X} : X_i \rhd X_k$, the second clause in the definition of $\mathcal{U} \succ_d \mathcal{V}$ trivially holds. Hence, $X_i$ is a witness for $\mathcal{U} \succ_d \mathcal{V}$ (see Definition 11). □

Note that the proof of the above proposition only made use of the fact that $\forall X_k \in \mathcal{X} : X_i \rhd X_k$, which is a weaker condition than $\rhd$ being a total order. Hence, we have the following more general result.

**Proposition 14.** *If $\rhd$ is such that there is a unique most important attribute $X_i$, i.e., $\exists X_i \in \mathcal{X} : \forall X_k \in \mathcal{X} \setminus \{X_i\} : X_i \rhd X_k$, then $\succ'_i \subseteq \succ_d$.*

We proceed to prove an important result that gives conditions under which $\succ_d$ is a weak order.

**Theorem 3.** *When the aggregation function $\succ'_i$ is defined as in Definition 8, if $\rhd$ as well as $\{\succ_i\}$ are total orders, then $\succ_d$ is a weak order.*

*Proof.* $\succ_d$ is a weak order if and only if it is a strict partial order and negatively transitive. We have already shown that $\succ_d$ is a strict partial order in Theorem 1, and hence we are only left with proving that $\succ_d$ is negatively transitive, i.e., $\mathcal{U} \not\succ_d \mathcal{V} \wedge \mathcal{V} \not\succ_d \mathcal{Z} \Rightarrow \mathcal{U} \not\succ_d \mathcal{Z}$.

First, we note that since $\succ_i$ is a total order, $\succ'_i$ is also a total order (see Definition 8).
$\mathcal{U} \not\succ_d \mathcal{V} \Rightarrow (\forall X_i : \mathcal{U}(X_i) \succ'_i \mathcal{V}(X_i) \Rightarrow \exists X_k : (X_k \rhd X_i \wedge \mathcal{U}(X_k) \not\succeq'_k \mathcal{V}(X_k)))$ ($X_k \sim_\rhd X_i$ is not possible because $\rhd$ is a total order). (1)

Let $X_i$ and $X_j$ be the most important attributes s.t. $\mathcal{U}(X_i) \succ'_i \mathcal{V}(X_i)$ and $\mathcal{V}(X_j) \succ'_j \mathcal{Z}(X_j)$ respectively. (2)

Let $X_p$ and $X_q$ be the most important attributes s.t. $X_p \rhd X_i \wedge \mathcal{U}(X_p) \not\succeq'_p \mathcal{V}(X_p)$ and $X_q \rhd X_j \wedge \mathcal{V}(X_q) \not\succeq'_q \mathcal{Z}(X_q)$ respectively (such $X_p$ and $X_q$ must exist by (1)). (3)

**Case 1** Both $X_i$ and $X_j$ as defined in (2) exist (cases when such $X_i$ and/or $X_j$ don't exist will be dealt with separately).

Three sub-cases arise: $X_p \rhd X_q$, $X_q \rhd X_p$ and $X_p = X_q$.
Case 1a: Suppose that $X_p \rhd X_q$ (see Figure 7). (4)

- From (3) we know that $X_p \rhd X_i \wedge \mathcal{U}(X_p) \not\succeq'_p \mathcal{V}(X_p)$, i.e., $\mathcal{V}(X_p) \succ'_p \mathcal{U}(X_p)$. (5)

- From (3) and (4) we know that $\mathcal{V}(X_p) \succeq'_p \mathcal{Z}(X_p)$, because $X_q$ is the most important attribute that is also more important than $X_j$ and $\mathcal{V}(X_q) \not\succeq'_q \mathcal{Z}(X_q)$, and $X_p$ is more important than $X_q$ (and hence $X_j$ as well). (6)

- But because $X_j$ is the most important attribute with $\mathcal{V}(X_j) \succ'_j \mathcal{Z}(X_j)$, and $X_p \rhd X_j$ (since $X_q \rhd X_j$ and $X_p \rhd X_q$), we have $\mathcal{V}(X_p) \not\succ'_p \mathcal{Z}(X_p)$ (as $X_j$ is the most important attribute with $\mathcal{V}(X_j) \succ'_j \mathcal{Z}(X_j)$, using (2)). Along with (6), this means that $\mathcal{V}(X_p) = \mathcal{Z}(X_p)$. (7)

- From (5) and (7), $\mathcal{Z}(X_p) \succ'_p \mathcal{U}(X_p)$. (8)

- Also, $\forall X_k : X_k \rhd X_p \Rightarrow \mathcal{U}(X_k) = \mathcal{V}(X_k) \wedge \mathcal{V}(X_k) = \mathcal{Z}(X_k)$ (because $X_k$ is more important than $X_i, X_j$ and $X_p, X_q$). (9)

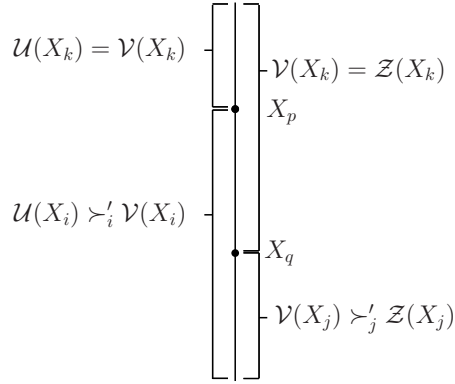

$\mathcal{U}(X_k) = \mathcal{V}(X_k)$

$\mathcal{V}(X_k) = \mathcal{Z}(X_k)$

$X_p$

$\mathcal{U}(X_i) \succ'_i \mathcal{V}(X_i)$

$X_q$

$\mathcal{V}(X_j) \succ'_j \mathcal{Z}(X_j)$

Figure 7: The case when $X_p \rhd X_q$

- From (8) and (9), $\mathcal{Z} \succ_d \mathcal{U}$ with $X_p$ as witness. Hence, $\mathcal{U} \not\succ_d \mathcal{Z}$.

Case 1b: Suppose that $X_q \rhd X_p$. The claim holds by symmetry.
Case 1c: Suppose that $X_p = X_q$.

- From (3) we know that $X_p \rhd X_i \wedge \mathcal{U}(X_p) \not\succeq'_p \mathcal{V}(X_p)$, i.e., $\mathcal{V}(X_p) \succ'_p \mathcal{U}(X_p)$.

- Similarly, $\mathcal{Z}(X_p) \succ'_p \mathcal{V}(X_p)$.

- Hence, $\mathcal{Z}(X_p) \succ'_p \mathcal{U}(X_p)$. Moreover, $\forall X_k : X_k \rhd X_p \Rightarrow \mathcal{U}(X_k) = \mathcal{V}(X_k) \wedge \mathcal{V}(X_k) = \mathcal{Z}(X_k)$ (because $X_k$ is more important than $X_i, X_j$ and $X_p, X_q$).

- Therefore, $\mathcal{Z} \succ_d \mathcal{U}$ with $X_p$ as witness. Hence, $\mathcal{U} \not\succ_d \mathcal{Z}$.

**Case 2** : If $X_i$ (say) does not exist, then $\forall X_i : \mathcal{U}(X_i) \not\succ'_i \mathcal{V}(X_i)$. Let $X_p$ be the most important attribute s.t. $\mathcal{V}(X_p) \succ'_p \mathcal{U}(X_p)$ (if $X_p$ does not exist, then trivially $\mathcal{U} \not\succ_d \mathcal{Z}$ because $\mathcal{U} = \mathcal{V}$). (10)

Case 2a: Suppose $X_p \rhd X_q$. Then $\forall X_k : X_k \rhd X_p \Rightarrow \mathcal{V}(X_k) = \mathcal{Z}(X_k)$ (because $X_k \rhd X_q$ as well). Moreover, $X_p \rhd X_q \Rightarrow \mathcal{V}(X_p) = \mathcal{Z}(X_p)$. Hence, $\mathcal{Z} \succ_d \mathcal{U}$ with $X_p$ as witness and therefore $\mathcal{U} \not\succ_d \mathcal{Z}$.

Case 2b: Suppose $X_q \rhd X_p$. Then $\forall X_k : X_k \rhd X_q \Rightarrow \mathcal{U}(X_k) = \mathcal{V}(X_k)$ (because $X_k \rhd X_p$ as well). Moreover, $X_q \rhd X_p \Rightarrow \mathcal{U}(X_q) = \mathcal{V}(X_q)$. Hence, $\mathcal{Z} \succ_d \mathcal{U}$ with $X_q$ as the witness and therefore $\mathcal{U} \not\succ_d \mathcal{Z}$.

Case 2c: Suppose $X_p = X_q$. Then $\forall X_k : X_k \rhd X_p \Rightarrow \mathcal{V}(X_k) = \mathcal{Z}(X_k)$ (because $X_k \rhd X_q$ as well) and similarly $\forall X_k : X_k \rhd X_q \Rightarrow \mathcal{U}(X_k) = \mathcal{V}(X_k)$ (because $X_k \rhd X_p$ as well). Moreover, since $\mathcal{V}(X_q) \not\succeq'_q \mathcal{Z}(X_q)$ (by (3)), $\mathcal{V}(X_p) \succ'_p \mathcal{U}(X_p)$ (using (10)) we have $\mathcal{Z}(X_p) \succ'_p \mathcal{U}(X_p)$. Hence, $\mathcal{Z} \succ_d \mathcal{U}$ with $X_p$ as the witness and therefore $\mathcal{U} \not\succ_d \mathcal{Z}$.

**Case 3** : If $X_j$ (say) does not exist, the proof is symmetric to Case 2.

**Case 4** : Suppose that both $X_i$ and $X_j$ do not exist. Then, for any attribute $X_i$, $\mathcal{V}(X_i) \succeq'_i \mathcal{U}(X_i)$ and $\mathcal{Z}(X_i) \succeq'_i \mathcal{V}(X_i)$), i.e., $\forall X_i : \mathcal{Z}(X_i) \succeq'_i \mathcal{U}(X_i)$. Hence, there is no witness for $\mathcal{U} \succ_d \mathcal{Z}$, or $\mathcal{U} \not\succ_d \mathcal{Z}$.

Cases 1 - 4 are exhaustive, and in each case $\mathcal{U} \not\succ_d \mathcal{Z}$. This completes the proof. $\qquad \square$

We further conjecture that $\succ_d$ is a weak order when $\{\succ_i\}$ are total orders and $\rhd$ is an arbitrary interval order (i.e., under conditions that are more general than the conditions of Theorem 3). We leave this as an open problem.

**Conjecture 1.** *If $\{\succ_i\}$ are total orders and $\rhd$ is an arbitrary interval order, then $\succ_d$ is a weak order.*

**Remark.**
As stated, Conjecture 1 and Theorem 3 apply whenever $\{\succ_i\}$ are totally ordered, and when using our method of comparing two aggregated valuations ($\succ'_i$) (see Definition 8). More generally, we note that they hold whenever $\{\succ'_i\}$ are total orders, regardless of the chosen method of comparing two aggregated valuations, and regardless of the properties of the input intra-attribute preferences $\{\succ_i\}$. For example, suppose that $\{\succ_i\}$ are ranked weak orders (i.e., not total orders). As such, Conjecture 1 and Theorem 3 do not apply. However, for each attribute $X_i$ if we define $\Phi_i(S)$ to be the rank number corresponding to the worst frontier of $S$, and $\succ'_i$ as the natural total order over the ranks in the weak order, then the consequences of Conjecture 1 and Theorem 3 hold.

We summarize the theoretical results relating the properties of the dominance relation and the properties of the preference relations $\rhd$ and $\{\succ'_i\}$ in Table 5.

| $\rhd$ | $\succ'_i$ | $\succ_d$ | Remarks |
|--------|-----------|-----------|--------------|
| *io* | *po* | *po* | Theorem 1 |
| *io* | *to* | *wo* | Conjecture 1 |
| *to* | *to* | *wo* | Theorem 3 |

Table 5: Summary of results and conjectures relating to the properties of $\succ_d$ with respect to the properties of $\rhd$ and $\{\succ'_i\}$.

## 3.7 Choosing the Most Preferred Solutions

Given a set $\mathfrak{C} = \{\mathcal{C}_i\}$ of compositions and a preference relation $\succ$ (e.g., $\succ_d$) that allows us to compare any pair of compositions, the problem is to find the most preferred composition(s). When the preference relations are totally ordered (e.g., a ranking) over a set of alternative solutions, rationality of choice suggests ordering the alternatives with respect to the complete preference and choosing the "*best*" alternative, i.e., the one that ranks the highest. However, when the preference relation is a strict partial order, e.g., in the case of $\succ_d$, not every pair of solutions (compositions) may be comparable. Therefore, a solution that is *the most preferred* with respect to the preference relation may not exist. Hence, we use the notion of the *non-dominated* set of solutions defined as follows.

**Definition 13** (Non-dominated Set). *The non-dominated set of elements (alternatives or solutions or compositions) of a set $\mathfrak{C}$ with respect to a (partially ordered) preference relation $\succ$ (e.g., $\succ_d$), denoted $\Psi_\succ(\mathfrak{C})$, is a subset of $\mathfrak{C}$ such that none of the elements in $S$ are preferred to any element in $\Psi_\succ(\mathfrak{C})$.*

$$\Psi_\succ(\mathfrak{C}) = \{\mathcal{C}_i \in \mathfrak{C} | \nexists \mathcal{C}_j \in \mathfrak{C} : \mathcal{C}_j \succ \mathcal{C}_i\}$$

Note that as per this definition, $\Psi_\succ(\mathfrak{C})$ is the *maximal* set of elements in $\mathfrak{C}$ with respect to the relation $\succ$. It is also easy to observe that $\mathfrak{C} \neq \emptyset \Leftrightarrow \Psi_\succ(\mathfrak{C}) \neq \emptyset$.

## 4. Algorithms for Computing the Most Preferred Compositions

We now turn to the problem of identifying from a set of feasible compositions (that satisfy a pre-specified functionality ($\varphi$)), the most preferred subset, i.e., the non-dominated set.

### 4.1 Computing the Maximal/Minimal Subset with Respect to a Partial Order

The straightforward way of computing the maximal (non-dominated) elements in a set $S$ of $n$ elements with respect to any preference relation $\succ$ is the following algorithm: For each element $s_i \in S$, check if $\exists s_j \in S : s_j \succ s_i$, and if not, $s_i$ is in the non-dominated set. This simple "compare all pairs and delete dominated" approach involves computing dominance with respect to $\succ$ $O(n^2)$ times.

Recently Daskalakis, Karp, Mossel, Riesenfeld and Verbin (2009) provided an algorithm that performs at most $O(wn)$ pairwise comparisons to compute the maximal elements of a set $S$ with respect to a partial order $\succ$, where $n = |S|$ and $w$ is the *width* of the partial order $\succ$ on $S$ (the size of the maximal set of pairwise incomparable elements in $S$ with respect to $\succ$). The algorithm presented by Daskalakis et al. finds the minimal elements; the corresponding algorithm for finding the maximal elements is as follows.
Let $T_0 = \emptyset$. Let the elements of the set $S$ be $x_1, x_2, \cdots x_n$. At step $t(\geq 1)$:

- Compare $x_t$ to all elements in $T_{t-1}$.

- If there exists some $a \in T_{t-1}$ such that $a \succ x_t$, do nothing.

- Otherwise, remove from $T_{t-1}$ all elements $a$ such that $x_t \succ a$ and put $x_t$ into $T_t$.

On termination, the set $T_n$ contains all the maximal elements in $S$, i.e., non-dominated subset of $S$ with respect to $\succ$. We make use of the above algorithm to compute the non-dominated (maximal) subsets (namely, $\Psi_\succ(\cdot)$), and the original version of the algorithm given in by Daskalakis (2009) to compute the worst-frontiers (minimal subsets).

### 4.2 Algorithms for Finding the Most Preferred Feasible Compositions

We proceed to develop algorithms for finding the most preferred feasible compositions, given a compositional system $\langle R, \oplus, \models \rangle$ consisting of a repository $R$ of pre-existing components, a user specified functionality $\varphi$, user preferences $\{\succ_i\}$ and $\triangleright$ and a functional composition algorithm $f$. We analyze the properties of the algorithms with respect to the worst-frontier based aggregation (see Definition 6).

**Definition 14** (Soundness and Completeness)**.** *An algorithm $A$ that, given a set $\mathfrak{C}$ of feasible compositions, computes a set of feasible compositions $S_A \subseteq \Psi_{\succ_d}(\mathfrak{C})$ is said to be sound with respect to $\mathfrak{C}$. Such an algorithm is complete with respect to $\mathfrak{C}$ if $S_A \supseteq \Psi_{\succ_d}(\mathfrak{C})$.*

---

**Algorithm 1** ComposeAndFilter($\succ, f, \varphi$)

---

  1. Find the set $\mathfrak{C}$ of feasible compositions w.r.t. $\varphi$ using $f$

  2. **return** $\Psi_\succ(\mathfrak{C})$

---

Given a compositional system $\langle R, \oplus, \models \rangle$ consisting of a repository $R$ of pre-existing components, and a user specified functionality $\varphi$, the most preferred approach to finding the most preferred feasible compositions involves: (a) computing the set $\mathfrak{C}$ of functionally feasible compositions using a functional composition algorithm $f$, and (b) choosing the non-dominated set according to preferences over non-functional attributes.

Algorithm 1 follows this simple approach to produce the set $\Psi_{\succ_d}(\mathfrak{C})$ of all non-dominated feasible compositions, when invoked with the preference relation $\succ_d$, the functional composition algorithm $f$ and the desired functionality $\varphi$. $\Psi_{\succ_d}(\mathfrak{C})$ can be computed using the procedure described in Section 4.1. Algorithm 1 is both sound and complete with respect to $\mathfrak{C}$.

### 4.3 A Sound and Weakly Complete Algorithm

Note that in the worst case, Algorithm 1 evaluates the dominance relation $\succ_d$ between all possible pairs of feasible compositions $\mathfrak{C}$. However, this can be avoided if we settle for a non-empty subset of $\Psi_{\succ_d}(\mathfrak{C})$. Note that every solution in such a subset is guaranteed to be "*optimal*" with respect to user preferences $\succ_d$. We introduce the notion of *weak completeness* to describe an algorithm that computes a set of feasible compositions, at least one of which is non-dominated with respect to $\succ_d$.

**Definition 15** (Weak Completeness)**.** *An algorithm $A$ that, given a set $\mathfrak{C}$ of feasible compositions, computes a set $S_A$ of feasible compositions is said to be weakly complete with respect to $\mathfrak{C}$ if $\Psi_{\succ_d}(\mathfrak{C}) \neq \emptyset \Rightarrow S_A \cap \Psi_{\succ_d}(\mathfrak{C}) \neq \emptyset$.*

We now proceed to describe a sound and weakly complete algorithm, i.e., one that computes a non-empty subset of $\Psi_{\succ_d}(\mathfrak{C})$. The algorithm is based on the following observation: Solutions that are non-dominated with respect to each of the relatively most-important attributes are guaranteed to include some solutions that are non-dominated overall with respect to $\succ_d$ as well. Hence, the solutions that are most preferred with respect to each such attribute can be used to compute a non-empty subset of $\Psi_{\succ_d}(\mathfrak{C})$. We proceed by considering solutions that are most preferred with respect to an attribute $X_i$.

**Definition 16** (Non-dominated solutions w.r.t. attributes)**.** *The set $\Psi_{\succ'_i}(\mathfrak{C})$ of solutions that are non-dominated with respect to an attribute $X_i$ is defined as*

$$\Psi_{\succ'_i}(\mathfrak{C}) = \{\mathcal{U} \mid \mathcal{U} \in \mathfrak{C} \wedge \nexists \mathcal{V} \in \mathfrak{C} : \mathcal{V}(X_i) \succ'_i \mathcal{U}(X_i)\}.$$

Let $I \subseteq \mathcal{X}$ be the set of most important attributes with respect to $\triangleright$, i.e., $I = \Psi_\triangleright(\mathcal{X}) = \{X_i \mid \nexists X_j \in \mathcal{X} : X_j \triangleright X_i\}$. Clearly, $I \neq \emptyset$ because there always exists a non-empty maximal set of elements in the partial order $\triangleright$. The following proposition states that for every $X_i \in I$, at least one of the solutions in $\Psi_{\succ'_i}(\mathfrak{C})$ is also contained in $\Psi_{\succ_d}(\mathfrak{C})$.

**Proposition 15.** *$\forall X_i \in I : \Psi_{\succ_d}(\mathfrak{C}) \neq \emptyset \Rightarrow \Psi_{\succ'_i}(\mathfrak{C}) \cap \Psi_{\succ_d}(\mathfrak{C}) \neq \emptyset$ (See Appendix A for a proof).*

Algorithm 2 constructs a subset of $\Psi_{\succ_d}(\mathfrak{C})$, using the sets $\{\Psi_{\succ_i'}(\mathfrak{C}) \mid X_i \in I\}$. First, the algorithm computes the set $I$ of most important attributes in $\mathcal{X}$ with respect to $\triangleright$ (Line 2). The algorithm iteratively computes $\Psi_{\succ_i'}(\mathfrak{C})$ for each $X_i \in I$ (Lines 3, 4), identifies the subset of solutions that are non-dominated with respect to $\succ_d$ in each case, and combines them to obtain $\theta \subseteq \Psi_{\succ_d}(\mathfrak{C})$.

---

**Algorithm 2** WeaklyCompleteCompose($\{\succ_i \mid X_i \in \mathcal{X}\}, \triangleright, f, \varphi$)

---

1. $\theta \leftarrow \emptyset$
2. $I \leftarrow \Psi_{\triangleright}(\mathcal{X}) = \{X_i \mid \nexists X_j : X_j \triangleright X_i\}$
3. **for all** $X_i \in I$ **do**
4.    $\Psi_{\succ_i'}(\mathfrak{C}) \leftarrow$ ComposeAndFilter($\succ_i', f, \varphi$)
5.    $\theta \leftarrow \theta \cup \Psi_{\succ_d}(\Psi_{\succ_i'}(\mathfrak{C}))$
6. **end for**
7. **return** $\theta$

---

**Theorem 4** (Soundness and Weak Completeness of Algorithm 2). *Given a set of attributes* $\mathcal{X}$, *preference relations* $\triangleright$ *and* $\succ_i'$, *Algorithm 2 generates a set* $\theta$ *of feasible compositions such that* $\theta \subseteq \Psi_{\succ_d}(\mathfrak{C})$ *and* $\Psi_{\succ_d}(\mathfrak{C}) \neq \emptyset \Rightarrow \theta \neq \emptyset$ *(See Appendix A for a proof).*

In general, Algorithm 2 is not guaranteed to yield a complete set of solutions, i.e., $\theta \neq \Psi_{\succ_d}(\mathfrak{C})$. The following example illustrates such a case.

***Example.*** Consider a compositional system with two attributes $\mathcal{X} = \{X_1, X_2\}$, with domains $\{a_1, a_2, a_3\}$ and $\{b_1, b_2, b_3\}$ respectively. Let their intra-attribute preferences be total orders: $a_1 \succ_1 a_2 \succ_1 a_3$ and $b_1 \succ_2 b_2 \succ_2 b_3$ respectively, and let both attributes be equally important ($\triangleright = \emptyset$). Suppose the user-specified goal $\varphi$ is satisfied by three feasible compositions $\mathcal{C}_1, \mathcal{C}_2, \mathcal{C}_3$ with valuations $V_{\mathcal{C}_1} = \langle \{a_1\}, \{b_3\}\rangle$, $V_{\mathcal{C}_2} = \langle \{a_3\}, \{b_1\}\rangle$ and $V_{\mathcal{C}_3} = \langle \{a_2\}, \{b_2\}\rangle$ respectively. Given the above preferences, $\Psi_{\succ_1'}(\mathfrak{C}) = \{\mathcal{C}_1\}$ and $\Psi_{\succ_2'}(\mathfrak{C}) = \{\mathcal{C}_2\}$. Thus, $\theta = \{\mathcal{C}_1, \mathcal{C}_2\}$ However, $\Psi_{\succ_d}(\mathfrak{C}) = \{\mathcal{C}_1, \mathcal{C}_2, \mathcal{C}_3\} \neq \theta$.    $\diamond$

The above example shows that the most preferred valuation for one attribute (e.g., $X_1$) can result in poor valuations for one or more other attributes (e.g., $X_2$). Algorithm 2 may thus leave out solutions like $\mathcal{C}_3$ that are not most preferred with respect to any one $\succ_i'$, but nevertheless may correspond to a good compromise when we consider multiple most important attributes. It is a natural question to ask what are the minimal conditions under which Algorithm 2 is complete. A related question is whether Algorithm 2 can be guaranteed to produce a certain minimum number of non-dominated solutions ($|\theta|$) under some specific set of conditions. Note that in general, the cardinality of $\theta$ depends not only on the user preferences $\succ_i, \triangleright$, but also on the user specified functionality $\varphi$ which together with the repository $R$ determines the set $\mathfrak{C}$ of feasible compositions. However, in the special case when $\triangleright$ specifies a single attribute $X_t$ that is relatively more important than all other attributes, we can show that Algorithm 2 is complete.

**Proposition 16.** *If* $I = \{X_t\} \wedge \forall X_k \neq X_t \in \mathcal{X} : X_t \triangleright X_k$, *then* $\Psi_{\succ_d}(\mathfrak{C}) \subseteq \theta$, *i.e., Algorithm 2 is complete (See Appendix A for a proof).*

It remains to be seen what are all the necessary and sufficient conditions for ensuring the completeness of Algorithm 2, and we plan to address this problem in the future.

### 4.4 Optimizing with Respect to One of the Most Important Attributes

As we will see in Section 5, Algorithm 2 has a high worst case complexity, especially if the set $I$ of most important attributes is large. This is due to the fact that for each most important attribute $X_i \in I$, the algorithm computes the non-dominated set over the feasible compositions with respect to $\succ_i'$ first, and then with respect to $\succ_d$, i.e., $\theta \cup \Psi_{\succ_d}(\Psi_{\succ_i'}(\mathfrak{C}))$ (Line 4). The computation of the non-dominated set with respect to $\succ_d$, although expensive, is crucial to ensuring the soundness of Algorithm 2.

While soundness is a desirable property, there may be settings requiring faster computation of feasible compositions, where it may be acceptable to obtain a set $S$ of feasible compositions that contains *at least one* (whenever there exists one) of the most preferred feasible compositions (one that is non-dominated by any other feasible composition with respect to $\succ_d$). In such a case, it might be useful to have an algorithm with lower complexity that finds a set of feasible compositions of which *at least one* is most preferred (i.e., weakly complete), as opposed to one with a higher complexity that finds a set of feasible compositions *all* of which are most preferred (i.e., sound).

---

**Algorithm 3** AttWeaklyCompleteCompose($\{\succ_i \mid X_i \in \mathcal{X}\}, \rhd, f, \varphi$)

---

1. $I \leftarrow \Psi_\rhd(\mathcal{X}) = \{X_i \mid \nexists X_j : X_j \rhd X_i\}$
2. **for some** $X_i \in I$
3. $\theta \leftarrow \Psi_{\succ_i'}(\mathfrak{C}) = \text{ComposeAndFilter}(\succ_i', f, \varphi)$
4. **return** $\theta$

---

We consider one such modification of Algorithm 2, namely Algorithm 3, that arbitrarily picks one of the most important attributes $X_i \in I$ (as opposed to the entire set $I$ as in Algorithm 2) and finds the set of all feasible compositions that are non-dominated with respect to $\succ_i'$, i.e., $\theta = \Psi_{\succ_i'}(\mathfrak{C})$ for Algorithm 3.

The weak completeness of Algorithm 3 follows directly from Proposition 15. In the following example, however, we show that some of the feasible compositions produced by Algorithm 3 may be dominated by some other feasible composition with respect to $\succ_d$, i.e., Algorithm 3 is not sound.

***Example.*** Consider a compositional system with two attributes $\mathcal{X} = \{X_1, X_2\}$, with domains $\{a_1, a_2\}$ and $\{b_1, b_2\}$ respectively. Let their intra-attribute preferences be: $a_1 \succ_1 a_2$ and $b_1 \succ_2 b_2$ respectively, and let both attributes be equally important ($\rhd = \emptyset$; $I = \{X_1, X_2\}$). Suppose the user-specified goal $\varphi$ is satisfied by three feasible compositions $\mathcal{C}_1, \mathcal{C}_2, \mathcal{C}_3$ with valuations $V_{\mathcal{C}_1} = \langle \{a_1\}, \{b_1\} \rangle$, $V_{\mathcal{C}_2} = \langle \{a_2\}, \{b_1\} \rangle$ and $V_{\mathcal{C}_3} = \langle \{a_1\}, \{b_2\} \rangle$ respectively. Given the above preferences, if we choose to maximize the preference with respect to attribute $X_1 \in I$, then $\theta = \Psi_{\succ_1'}(\mathfrak{C}) = \{\mathcal{C}_1, \mathcal{C}_3\}$. If we choose $X_2 \in I$ instead, we get $\theta = \Psi_{\succ_2'}(\mathfrak{C}) = \{\mathcal{C}_1, \mathcal{C}_2\}$. However, in any case $\Psi_{\succ_d}(\mathfrak{C}) = \{\mathcal{C}_1\} \neq \theta$. ◇

The following proposition gives a condition under which Algorithm 3 is complete.

**Proposition 17.** *If $|I| = 1$, i.e., there is a unique most important attribute with respect to $\rhd$, then Algorithm 3 is complete (See Appendix A for a proof).*

---

**Algorithm 4** $InterleaveCompose(\mathcal{L}, \succ, f, \varphi)$

---

1. **if** $\mathcal{L} = \emptyset$ **then**
2.      **return** $\emptyset$
3. **end if**
4. $\theta = \Psi_\succ(\mathcal{L})$
5. $\theta' = \emptyset$
6. **for all** $\mathcal{C} \in \theta$ **do**
7.      **if** $\mathcal{C} \not\models \varphi$ **then**
8.          $\theta' = \theta' \cup f(\mathcal{C})$
9.      **else**
10.          $\theta' = \theta' \cup \{\mathcal{C}\}$
11.      **end if**
12. **end for**
13. **if** $\theta' = \theta$ **then**
14.      **return** $\theta$
15. **else**
16.      $InterleaveCompose((\mathcal{L} \setminus \theta) \cup \theta', \succ, f, \varphi)$
17. **end if**

---

## 4.5 Interleaving Functional Composition with Preferential Optimization

Algorithms 1, 2 and 3 identify the most preferred feasible compositions using the two step approach: (a) find the feasible compositions $\mathfrak{C}$; and (b) compute a subset of $\mathfrak{C}$ that is preferred with respect to the user preferences. We now develop an algorithm that eliminates some of the intermediate partial feasible compositions from consideration based on the user preferences. This is particularly useful in settings (such as when $|\mathfrak{C}|$ is large relative to $|\Psi_{\succ_d}(\mathfrak{C})|$), where it might be more efficient to compute only a subset of $\mathfrak{C}$ that are likely (based on $\succ_i$ and $\rhd$) to be in $\Psi_{\succ_d}(\mathfrak{C})$.

Algorithm 4 requires that the functional composition algorithm $f$ is incremental (see Definition 2), i.e., that it produces a set $f(\mathcal{C})$ of functionally feasible extensions given any existing partial feasible composition $\mathcal{C}$. At each step, Algorithm 4 chooses a subset of the feasible extensions produced by applying $f$ on all the non-dominated partial feasible compositions, based on the user preferences. Algorithm 4 computes the non-dominated set of feasible compositions by *interleaving* the execution of the incremental functional composition algorithm $f$ with the ordering of partial solutions with respect to preferences over non-functional attributes.

Algorithm 4 is initially invoked using the parameters $\mathcal{L} = (\bot)$ [11], $\succ_d$, the functional composition algorithm $f$ and $\varphi$. The algorithm maintains at each step a list $\mathcal{L}$ of partial feasible compositions under consideration. If $\mathcal{L}$ is empty at any step, i.e., there are no more partial feasible compositions to be explored, then the algorithm terminates with no solution (Lines $1-3$); otherwise it selects from $\mathcal{L}$, the subset $\theta$ that is non-dominated with respect to some preference relation $\succ$ (Line 4). If all the partial feasible compositions in $\theta$ are also feasible compositions, then the algorithm outputs $\theta$ and terminates (Lines $13-14$). Otherwise, it replaces the partial feasible compositions in $\theta$ that are not feasible compositions, with their one-step extensions (Lines $7-8$). The algorithm continues to recurse (Line 16), at each iteration updating the dominated set by replacing $\theta$ with $\theta'$ until there are no changes in the dominated set i.e., $\theta = \theta'$. Note that it is not possible to eliminate the dominated compositions ($\mathcal{L} \setminus \theta$) at this stage because some of their extensions (in a later iteration) could result in non-dominated compositions.

---

11. It is not necessary to invoke the algorithm with $\mathcal{L} = (\bot)$ (i.e., only $\bot$ in the list $\mathcal{L}$) initially. There may be functional composition algorithms that begin with an non-empty composition $\mathcal{C}$ and proceed to obtain a feasible composition by iteratively altering $\mathcal{C}$. For instance, one could think of randomized or evolutionary algorithms that begin with a random, non-empty composition which is somehow repeatedly "improved" during the course of composition.

**Proposition 18** (Termination of Algorithm 4). *Given a finite repository of components, Algorithm 4 terminates in a finite number of steps (See Appendix A for a proof).*

We next investigate the soundness, weak-completeness and completeness properties of Algorithm 4. Proposition 19 states that the algorithm is in general not sound with respect to $\mathfrak{C}$, i.e., it is not guaranteed to produce feasible compositions that are non-dominated with respect to $\succ_d$. However, this does not discount the usefulness of the algorithm, as we will show that it is sound under some other assumptions (see Theorem 5).

**Proposition 19** (Unsoundness of Algorithm 4). *Given a functional composition algorithm $f$ and user preferences $\succ'_i$ and $\triangleright$ over a set of attributes $\mathcal{X}$, Algorithm 4 is not guaranteed to generate a set of feasible compositions $\theta$ such that $\theta \subseteq \Psi_{\succ_d}(\mathfrak{C})$ (See Appendix A for a proof).*

This result implies that in general, not all feasible compositions returned by Algorithm 4 ($\theta$) are in $\Psi_{\succ_d}(\mathfrak{C})$. The example shown in Figure 8 illustrates this problem. At the time of termination, there may exist some partial feasible composition $\mathcal{B}$ in the list $\mathcal{L}$ that *is dominated* by some feasible composition $\mathcal{E}$ in $\theta$; however, it may be possible to extend $\mathcal{B}$ to a feasible composition $\mathcal{B} \oplus W$ that *dominates* one of the compositions $\mathcal{F}$ in $\theta$ (as illustrated by the counter example in the proof, see Appendix A). In other words, $V_{\mathcal{E}} \succ_d V_{\mathcal{B}}$, $V_{\mathcal{F}} \sim_d V_{\mathcal{E}}$, $V_{\mathcal{F}} \sim_d V_{\mathcal{B}}$ and $V_{\mathcal{B} \oplus W} \succ_d V_{\mathcal{F}}$.

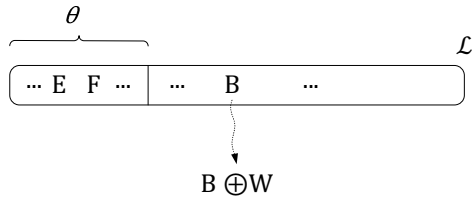

Figure 8: The case when Algorithm 4 is not sound

Although Algorithm 4 is not sound in general, we show that it is sound when the $\succ_d$ relation is an interval order (as opposed to an arbitrary partial order).

**Theorem 5** (Soundness of Algorithm 4). *If $\succ_d$ is an interval order, then given a functional composition algorithm $f$ and user preferences $\{\succ'_i\}, \triangleright$ over a set of attributes $\mathcal{X}$,*

*Algorithm 4 generates a set $\theta$ of feasible compositions such that $\theta \subseteq \Psi_{\succ_d}(\mathfrak{C})$ (See Appendix A for a proof).*

Because Theorem 5 requires $\succ_d$ to be an interval order, an important question arises: What are the conditions under which $\succ_d$ an interval order? Theorem 3 (see Section 3.6) gives us one such condition when $\succ_d$ is a weak order (i.e., also an interval order). The next two theorems give conditions under which Algorithm 4 is weakly complete and complete respectively.

**Theorem 6** (Weak Completeness of Algorithm 4). *If $\succ_d$ is an interval order, then given a functional composition algorithm $f$ and user preferences $\{\succ'_i\}, \triangleright$ over a set of attributes $\mathcal{X}$, Algorithm 4 produces a set $\theta$ of feasible compositions such that $\Psi_{\succ_d}(\mathfrak{C}) \neq \emptyset \Rightarrow \theta \cap \Psi_{\succ_d}(\mathfrak{C}) \neq \emptyset$ (See Appendix A for a proof).*

**Theorem 7** (Completeness of Algorithm 4). *If $\succ_d$ is a weak order, then given a functional composition algorithm $f$ and user preferences $\{\succ'_i\}, \triangleright$ over a set of attributes $\mathcal{X}$, Algorithm 4 generates a set $\theta$ of feasible compositions such that $\Psi_{\succ_d}(\mathfrak{C}) \subseteq \theta$ (See Appendix A for a proof).*

**Remark.** The above algorithm does not explore feasible compositions that can be generated by extending other feasible compositions (by the condition in Line 7). Proposition 6 shows that when worst-frontier based aggregation is used, extending a feasible composition cannot yield a more preferred feasible composition. This guarantees the soundness of Algorithm 4 (Theorem 5). However, when other aggregation schemes are used, it might be the case that a feasible extension of a feasible composition is more preferred, in which case, in order to ensure the soundness of Algorithm 4, Line 10 will have to be changed to $\theta' = \theta' \cup \{\mathcal{C}\} \cup f(\mathcal{C})$.

A summary of the conditions (in terms of the properties of the relative importance or dominance preference) under which the algorithms are sound, complete and weak complete are given in Table 6.

| Algorithm | Sound | Weakly Complete | Complete |
|-----------|-------|-----------------|----------|
| $A1$ | $po$ | $po$ | $po$ |
| $A2$ | $po$ | $po$ | $|I| = 1$ |
| $A3$ | $-$ | $po$ | $|I| = 1$ |
| $A4$ | $io$ | $io$ | $wo$ |

Table 6: Properties of $\succ_d$ or $\triangleright$ for which the algorithms are sound, weakly complete and complete. *po* stands for $\succ_d$ being a partial order; *io* stands for $\succ_d$ being an interval order; *wo* stands for $\succ_d$ being a weak order; and $|I| = 1$ is when $\triangleright$ is such that there is a unique most important attribute. '$-$' indicates that condition(s) under which $A3$ is sound remains as an open problem.

## 5. Complexity

In this section, we study the complexity of dominance testing (evaluating $\succ_d$, see Section 3.3) as well as the complexity of the algorithms for computing the non-dominated set of feasible compositions (see Section 4). We express the worst case time complexity of dominance testing in terms of the size of user specified intra-attribute, relative importance preference relations and the attribute domains (see Table 7).

| Relation / Set | Symbol | Cardinality | Remarks |
|---|---|---|---|
| Attributes | $\mathcal{X}$ | $m$ | Number of attributes |
| Domain of Attributes | $D_i$ | $n$ | Number of possible valuations of $X_i$ |
| Intra-attribute preferences | $\succ_i$ | $w_{int}$ | Width of the partial order $\succ_i$ |
| Intra-attribute preferences | $\succ_i$ | $k_{int}$ | Size of the relation $\succ_i$ |
| Relative Importance | $\rhd$ | $w_{rel}$ | Width of the partial order $\rhd$ |
| Relative Importance | $\rhd$ | $k_{rel}$ | Size of the relation $\rhd$ |
| Most Important Attributes | $I$ | $m_I$ | Number of most important attributes |
| Repository | $R$ | $r$ | Number of components in $R$ |
| Feasible Compositions | $\mathfrak{C}$ | $c$ | Number of feasible compositions |
| Dominance Relation | $\succ_d$ | $w_{dom}$ | Width of the dominance relation |

Table 7: Cardinalities of sets and relations

### 5.1 Computing the Maximal(Non-dominated)/Minimal Subset.

Let $\succ$ be a partial order on the set $S$, with a width of $w$ (size of the maximal set of elements which are pairwise incomparable) and $n = |S|$. The algorithm due to Daskalakis et al. discussed in Section 4.1 finds the maximal or minimal subset of $S$ with respect to $\succ$ within $O(wn)$ pairwise comparisons. Note that the maximum width of any partial order is $w = n$, when $\succ = \emptyset$. Hence, in the worst case $O(n^2)$ comparisons are needed.

### 5.2 Complexity of Dominance Testing

**Computing Worst Frontiers ($\Phi_i$).**    Let $S \subseteq D_i$. Recall from Definition 6 that the worst frontier of a set $S$ with respect to an attribute $X_i$ is $\Phi_i(S) := \{v : v \in S, \nexists u \in S \ s.t. \ v \succ_i u\}$, i.e., the minimal set of elements in $S$ with respect to the preference relation $\succ_i$. Using the algorithm due to Daskalakis et al. to find the minimal set with respect to a partial order (see above), the complexity of computing $\Phi_i(S)$ is $O(nw_{int})$.

**Comparing Worst Frontiers ($\succ'_i$).**    Let $A, B \in \mathscr{F}(X_i)$. As per Definition 8, $A \succ'_i B \Leftrightarrow \forall b \in B, \exists a \in A : a \succ_i b$. In the worst case, computing $A \succ'_i B$ would involve checking whether $a \succ_i b$ for each pair $a, b$, which would cost $O(k_{int})$. Hence, the complexity of comparing the worst frontiers $A$ and $B$ is $O(n^2 k_{int})$.

**Dominance Testing ($\succ_d$).**    Recall from Definition 11 the definition of dominance:

$$\mathcal{U} \succ_d \mathcal{V} \Leftrightarrow \quad \exists X_i \ : \ \mathcal{U}(X_i) \succ'_i \mathcal{V}(X_i) \ \wedge$$
$$\forall X_k, (X_k \rhd X_i \vee X_k \sim_\rhd X_i) \ \Rightarrow \ \mathcal{U}(X_k) \succeq'_k \mathcal{V}(X_k)$$

The complexity of dominance testing is the complexity of finding a witness attribute in $\mathcal{X}$ for $\mathcal{U} \succ_d \mathcal{V}$. For each attribute $X_i$, the complexity of computing the first clause in the conjunction of the definition of $\mathcal{U} \succ_d \mathcal{V}$ is $O(n^2 k_{int})$; and that of computing the second clause is $O\big(m(n^2 k_{int} + k_{rel})\big)$, where $O(k_{rel})$ and $O(n^2 k_{int})$ are the complexities of evaluating the left and right hand sides of the implication (respectively) for each $X_k \in \mathcal{X}$. Hence, the complexity of dominance testing is $O\big(m(n^2 k_{int} + m(n^2 k_{int} + k_{rel}))\big)$, or simply $O\big(m^2(n^2 k_{int} + k_{rel})\big)$. We will use the shorthand $d$ to denote $m^2(n^2 k_{int} + k_{rel})$.

## 5.3 Complexity of Algorithms

Each of the algorithms for computing the non-dominated feasible compositions (presented in Section 4) makes use of a functional composition algorithm $f$ to find feasible compositions. Hence, the complexity analysis of the algorithms needs to incorporate of the complexity of the functional composition algorithm as well.

Recall that Algorithms 1, 2 and 3 begin by computing the set of all feasible compositions in a *single shot* using a functional composition algorithm as a *black box*, and then proceed to find the most preferred among them. Algorithm 4 instead makes use of a functional composition algorithm that produces the set of feasible compositions by iteratively extending partial feasible compositions. Specifically, it *interleaves* the execution of the functional composition algorithm with the ordering of partial solutions with respect to preferences over non-functional attributes.

We denote by $O(f_e)$ and $O(f_g)$ respectively, the complexity of computing the set of feasible extensions of a partial feasible composition with respect to $\varphi$ and the complexity of computing the set of all feasible compositions with respect to $\varphi$.

## 5.4 Complexity of Algorithm 1

The overall complexity for finding the set of all non-dominated feasible compositions is $O(f_g + cw_{dom}d)$, where $O(d)$ is the complexity of evaluating $\succ_d$ for any pair of compositions. The first term $f_g$ accounts for Line 1 of the algorithm which computes the set of all feasible compositions, and the term $cw_{dom}d$ corresponds to the computation of $\Psi_{\succ_d}(\mathfrak{C})$ as per the algorithm given in Section 4.1.

## 5.5 Complexity of Algorithm 2

The complexity of identifying the most important attributes $I$ with respect to $\rhd$ (Line 1) is $O(mw_{rel}k_{rel})$. For *each* most important attribute $X_i \in I$, Algorithm 2 (a) invokes Algorithm 1 using the derived intra-attribute preference $\succ'_i$ to compute $\Psi_{\succ'_i}(\mathfrak{C})$; (b) identifies the subset of $\Psi_{\succ'_i}(\mathfrak{C})$ that is non-dominated with respect to $\succ_d$; and (c) adds them to the set of solutions. Hence, the complexity of Algorithm 2 is $O\big(mw_{rel}k_{rel} + m_I(f_g + cw_{dom}n^2 k_{int}) + m_I|\Psi_{\succ'_i}(\mathfrak{C})|^2 d\big)$.

Since the feasible compositions with respect to any given $\varphi$ are fixed, by computing the feasible compositions only once (during the first invocation of Algorithm 1 and storing them), the complexity of Algorithm 2 can be further reduced to $O(f_g + mw_{rel}k_{rel} + m_I cw_{dom}n^2 k_{int} + m_I|\Psi_{\succ'_i}(\mathfrak{C})|^2 d)$.

## 5.6 Complexity of Algorithm 3

The complexity of identifying the most important attributes $I$ with respect to $\triangleright$ (Line 1) is $O(mw_{rel}k_{rel})$. In contrast to Algorithm 2, Algorithm 3 invokes Algorithm 1 using the derived intra-attribute preference $\succ'_i$ to compute $\Psi_{\succ'_i}(\mathfrak{C})$ for *exactly one* of the most important attributes, $X_i \in I$. Hence, the complexity of Algorithm 3 is $O\big(f_g + mw_{rel}k_{rel} + cw_{dom}n^2k_{int}\big)$.

## 5.7 Complexity of Algorithm 4

We consider the worst case wherein the space of partial feasible compositions explored by Algorithm 4 is a tree rooted at $\bot$; let $b$ be its maximum branching factor (corresponding to the maximum number of extensions produced by the functional composition algorithm), and $h$ its height (corresponding to the maximum number of components used in a composition that satisfies $\varphi$). In the worst case, in each iteration of Algorithm 4, every element of $\mathcal{L}$, the list of current partial feasible compositions, ends up in the non-dominated set $\theta$.

Each level in the tree corresponds to one iteration of Algorithm 4, and at the $l$th iteration, in the worst case there are $b^l$ nodes in $\mathcal{L}$. Hence, the complexity of the $l$th iteration is $O\big((b^l)^2d + b^lf_e\big)$, where the first term corresponds to computing the non-dominated set among the current set of partial feasible compositions, and the second term corresponds to computing the feasible extensions of each partial feasible composition. Hence, the overall complexity of Algorithm 4 is $O\big(\sum_{l=0}^{h}(b^{2l}d + b^lf_e)\big) \approx O(b^{2h}d + b^hf_e)$.

We further conducted experiments on the algorithms using simulated problem instances to study how the algorithms perform in practice, which we describe next.

# 6. Experiments, Results & Analysis

We now describe the design and results of our experiments aimed at comparing the algorithms described in Section 4 with respect to the following attributes.

a) **Quality of solutions produced by the algorithms.** We measure the quality of the solutions produced by the algorithms as follows. First, among all the most preferred solutions that exist to the composition problem, we measure the fraction that is produced by the algorithm. Second, among all the solutions produced by the algorithm, we measure the fraction of solutions that are most preferred for the composition problem.

b) **Performance and efficiency of the algorithms.** The performance of an algorithm is measured in terms of response time (time taken to return the set of solutions), and the efficiency is measured in terms of the number of times an algorithm invokes the functional composition algorithm.

## 6.1 Experimental Setup

We now describe the data structure used to model the search space of compositions and the simulation parameters used to generate the compositions in our experiments.

### 6.1.1 MODELING THE SEARCH SPACE OF COMPOSITIONS USING RECURSIVE TREES

The *uniform recursive tree* (Smythe & Mahmoud, 1995) serves as a good choice to model the search space of partial compositions and their feasible extensions. A tree with $n$ vertices labeled by $1, 2, \ldots n$ is a *recursive tree* if the node labeled 1 is distinguished as the root, and $\forall k : 2 \leq k \leq n$, the labels of the nodes in the unique path from the root to the node labeled with $k$ form an increasing sequence. A *uniform recursive tree* of $n$ nodes (denoted $URTree(n)$) is one that is chosen with equal probability from the space of all such trees.

A simple growth rule can be used to generate a uniform random recursive tree of $n$ nodes, given such a tree of $n-1$ nodes: Given $URTree(n-1)$, choose uniformly at random a node in $URTree(n-1)$, and add a node labeled $n$ with the randomly chosen node as parent to obtain $URTree(n)$. The properties of this class of uniform random recursive trees are well studied in the literature of random data structures (see Smythe & Mahmoud, 1995, for a survey).

The rationale behind choosing the uniform recursive tree data structure to model the search space of our problem is that the growth rule that generates the recursive tree is similar in intuition to the process of searching for a feasible composition. Recall that the search space of partial compositions is generated by the recursive application of the functional composition algorithm $f$. The nodes in the recursive tree correspond to components in the repository of the composition problem. The tree is built starting with the root node – the search for feasible compositions correspondingly begins with $\bot$. The recursive tree is further grown by attaching new nodes to any of the existing nodes – this corresponds to extending feasible partial compositions by adding (composing) new components to any of the existing feasible partial compositions. Finally, the leaves of a recursive tree at depth $d$ from the root correspond to a (possibly feasible) composition of $d$ components from the repository in the composition problem.

We now show the precise correspondence between a recursive tree data structure and a search space of partial compositions.

- Each node in the tree corresponds to a composition.

- The root node corresponds to the empty composition $\bot$,

- Each node at level 1 corresponds to the composition of $\bot$ with a component $W$ in the repository, i.e., $\bot \oplus W = W, W \in R$,

- Each node at level $i$ corresponds to the composition of a component $W$ in the repository with the composition associated with the parent of this node,

- A leaf node is called a *feasible node* if the composition associated with this node satisfies $\varphi$.

For the purpose of experimentally evaluating our algorithms for finding the most preferred compositions and to compare them, we generate random recursive trees with varying number of nodes (or $|R|$, the number of components in the repository). In the generated random recursive tree, a certain fraction ($feas$) of leaves are picked uniformly randomly and are labeled to be feasible compositions. For each node in the generated and labeled random recursive tree, the valuation of attributes $\mathcal{X} = \{X_i\}$ (corresponding to the partial composition it represents) is randomly generated based on the respective domains $(\{D_i\})$[12].

### 6.1.2 User Preferences

We generate user preferences by generating random partial/total orders for each $\succ_i$ and random interval/total order for $\rhd$ for varying number of attributes $m = |\mathcal{X}|$ and domain size of attributes $n = |D_i|$.

A summary of the simulation parameters is given in Table 8.

| Parameter | Meaning | Range |
|---|---|---|
| $feas$ | Fraction of leaves in the search tree that are feasible compositions | $\{0.25, 0.5, 0.75, 1.0\}$ |
| $|D_i|$ | Domain size of preference attributes | $\{2, 4, 6, 8, 10\}$ |
| $|\mathcal{X}|$ | Number of preference attributes | $\{2, 4, 6, \ldots 20\}$ |
| $|R|$ | Number of components in the repository (nodes in the search tree) | $\{10, 20, \ldots 200\}$ |
| $fdelay$ | Overhead (in milliseconds) per invocation of the step-by-step functional composition algorithm $f$ | $\{1, 10, 100, 1000\}$ |
| $\succ_i$ | Intra-attribute preference over the values of $X_i$ | $\{po, to\}$ |
| $\rhd$ | Relative importance preference over $\mathcal{X}$ | $\{io, to\}$ |

Table 8: Simulation parameters and their ranges

### 6.1.3 Implementation of Algorithms

**Computing Dominance**

In order to check if one valuation dominates another with respect to the user preferences $\{\succ_i\}$ and $\rhd$, we iterate through all attributes $\mathcal{X}$ and check if there exists a witness for the dominance to hold (see Definition 11).

**Computing the most preferred solutions**

We implemented algorithms $A1$, $A2$, $A3$ and $A4$ in Java. Preliminary experiments with $A2$ showed that the algorithm did not scale up for large problem instances. In particular, when the number of attributes is large and dominance testing is computationally intensive, $A2$ timed out due to the computation of the non-dominated set multiple times for each of the most important attributes. Hence we did not proceed to run experiments on the samples with $A2$. However, we were able to run experiments with algorithm $A3$ that arbitrarily picks *one* of the most important attributes and finds the most preferred solutions with respect to the intra-attribute preferences of that attribute.

In algorithms $A1$ and $A3$ we first compute all solutions using the functional composition algorithm (simulated by $f$), whereas in $A4$, we interleave calls to $f$ with choosing preferred compositions (partial solutions) at each step. At each step, $A4$ chooses a subset of the feasible extensions of the current compositions for further exploration. Table 9 gives a brief description of the implemented algorithms.

---

12. Note that in the setup described here, the valuations for attributes is generated randomly for each node. In real applications, the valuations of the nodes may depend on the valuations of their parents.

| Alg. | Name of Algorithm | Remarks |
|---|---|---|
| $A1$ | *ComposeAndFilter* | First identifies functionally feasible compositions; then finds the non-dominated set of feasible compositions with respect to $\succ_d$ |
| $A2$ | *WeaklyCompleteCompose* | First identifies functionally feasible compositions; then finds the non-dominated set of feasible compositions with respect to $\succ_i$ for the most important attributes $\{X_i\}$ |
| $A3$ | *AttWeaklyCompleteCompose* | First identifies functionally feasible compositions; then picks an arbitrary most important attribute $X_i$ and finds the non-dominated set of feasible compositions with respect to $\succ_i'$ |
| $A4$ | *InterleaveCompose* | Identifies the non-dominated set of feasible extensions with respect to $\succ_d$ at each step; and recursively identifies their feasible extensions until all the non-dominated feasible extensions are feasible compositions |

Table 9: Implemented Algorithms

Table 10 shows the attributes that are recorded during the execution of each of the algorithms $A1, A3$ and $A4$ for each composition problem.

## 6.2 Results

We compare the algorithms $A1, A3, A4$ with respect to:

1. Quality of solutions produced by the algorithms, in terms of $SP/PF$ and $SP/S$

2. Performance and efficiency in terms of running time and number of calls to the functional composition algorithm $f$

### 6.2.1 QUALITY OF SOLUTIONS

We compare the quality of solutions produced by the algorithms in terms of the following measures.

| Attribute | Meaning | Remarks |
|-----------|---------|---------|
| $F$ | Set of solutions (feasible compositions) in a sample problem instance | $F = \mathfrak{C}_\varphi$ |
| $PF$ | Set of most preferred solutions in a sample problem instance with respect to the user preferences and the dominance relation | $PF = \Psi_{\succ_d}(\mathfrak{C}_\varphi) \subseteq F$ |
| $S$ | Set of solutions produced by the composition algorithm | |
| $SP$ | Set of solutions produced by the composition algorithm that are also most preferred solutions with respect to the user preferences and the dominance relation | $SP = PF \cap S \subseteq S$ |
| $T$ | Running time of the composition algorithm (ms) | |
| $fcount$ | Number of times the algorithm invokes the step-by-step functional composition algorithm $f$ | |

Table 10: Attributes observed during the execution of each algorithm

- $SP/PF$[13]: Proportion of most preferred solutions produced by the algorithm (fraction of all optimal solutions produced by the algorithm). If the algorithm is complete, then $SP/PF = 1$.

- $SP/S$: Proportion of solutions produced by the algorithm that are most preferred (fraction of solutions produced by the algorithm, that are optimal). If the algorithm is sound, then $SP/S = 1$.

The algorithm $A1$ exhaustively searches the entire space of compositions to identify all the feasible compositions $F$, and then finds the most preferred among them with respect to the user preferences $\rhd$ and $\{\succ_i\}$. Because it computes the set $\Psi_{\succ_d}(F)$, we observed that for $A1$, $SP = PF = S$, i.e., it is both sound (finds only the most preferred solutions) and complete (finds all the most preferred solutions).

We next compare the algorithms $A3$ and $A4$ with respect to $SP/PF$ and $SP/S$ for various types of ordering restrictions on the user preferences $\{\succ_i\}$ and $\rhd$. Table 11 reports results for the following combinations: (i) $\rhd$ is an interval order, $\{\succ_i\}$ are partial orders; (ii) $\rhd$ is an interval order, $\{\succ_i\}$ are total orders; (iii) $\rhd$ is a total order, $\{\succ_i\}$ are partial orders; and (iv) $\rhd$ is a total order, $\{\succ_i\}$ are total orders.

**Comparison of $SP/PF$**

- In general, most of the most preferred solutions were found by both the algorithms (see Table 11).

---

13. For the sake of readability, we use the notation used to denote the set to denote its cardinality as well, e.g., $SP$ is used to denote both the set and its cardinality ($|SP|$).

| $\triangleright$ | $\succ_i$ | $A3$ | $A4$ |
|---|---|---|---|
| $io$ | $po$ | 77.50 | 83.95 |
| $io$ | $to$ | 71.00 | 100.00 |
| $to$ | $po$ | 100.00 | 85.88 |
| $to$ | $to$ | 100.00 | 100.00 |

Table 11: Comparison of $SP/PF$ for algorithms $A3$ and $A4$ with respect to various ordering restrictions on $\{\succ_i\}, \triangleright$. The percent of problem instances for which $SP/PF = 1$ is shown in each row with respect to the corresponding ordering restrictions on the preference relations $\triangleright$ and $\{\succ_i\}$. The parameters used for simulating the problem instances and their ranges are given in Table 8.

- We observe that when relative importance ($\triangleright$) is a total order and $\{\succ_i\}$ are arbitrary partial orders, 100% of the most preferred solutions are produced by $A3$. Propositions 13 and 14 (see Section 3.6) were obtained based on this insight.

| $\triangleright$ | $\succ_i$ | $A3$ | $A4$ |
|---|---|---|---|
| $io$ | $po$ | 41.78 | 98.45 |
| $io$ | $to$ | 30.78 | 100.00 |
| $to$ | $po$ | 33.90 | 96.98 |
| $to$ | $to$ | 27.30 | 100.00 |

Table 12: Comparison of $SP/S$ for algorithms $A3$ and $A4$ with respect to various ordering restrictions on $\{\succ_i\}, \triangleright$. The percent of problem instances for which $SP/S = 1$ is shown in each row with respect to the corresponding ordering restrictions on the preference relations $\triangleright$ and $\{\succ_i\}$. The parameters used for simulating the problem instances and their ranges are given in Table 8.

**Comparison of $SP/S$**

- In general, most of the solutions that were found by the interleaved algorithm $A4$ were the most preferred solutions (see Table 12). On the other hand, algorithm $A3$ produced many solutions that were not the most preferred.

- The second (and fourth) row(s) of Tables 12 and 11 suggests that when intra-attribute preferences ($\{\succ_i\}$) are total orders and $\triangleright$ is an arbitrary interval order, the interleaved algorithm $A4$ is sound and complete, i.e., it produces exactly the non-dominated set of solutions with respect to $\succ_d$. Conjecture 1 and Theorem 3 in Section 3.6 were obtained based on this insight.

### 6.2.2 Performance and Efficiency

We compare the performance and efficiency of $A3, A4$ in terms of the number of times the functional composition algorithm $f$ is invoked, and running time (in milliseconds) for the algorithms to compute their solutions.

**Number of calls to functional composition $f$**

The plots in Figures 9 and 10 show the results of our experiments performed on problem instances where relative importance preferences are interval/total orders and intra-attribute preferences are partial/total orders, and they yield the following observations.

- In general, our experiments show that the interleaved algorithm $A4$ makes fewer calls to $f$ compared to $A3$. This can be seen in Figures 9 and 10, where all the data points corresponding to the number of calls to $f$ made by $A4$ (colored red) lie below those that correspond to $A3$ (colored green) in plots (a) and (b). This is because $A4$ explores only the most preferred subset of the available feasible extensions at each step in the search. On the other hand, $A3$ exhaustively explores all feasible extensions at each step.

- When the intra-attribute preferences $\{\succ_i\}$ are total orders, the difference in the number of calls to $f$ made by $A3$ and $A4$ is more pronounced. This can be observed in Figures 9 and 10, where the data points corresponding to the number of calls to $f$ made by $A4$ (colored red) lie much closer to the axis corresponding to the number of feasible compositions, in comparison to $A3$ (colored green). This can be explained by the fact that in this case the dominance relation is larger, due to which the number of incomparable pairs of compositions is smaller. Therefore, at each interleaving step the non-dominated set computed for extension is smaller.

- For both $A3$ and $A4$, the number of calls to $f$ decreases as the fraction of feasible compositions ($feas$) increases. Figures 9 and 10 show that as the number of feasible compositions increases, the data points corresponding to the number of calls to $f$ (for all algorithms) gets closer to the axis corresponding to the number of feasible compositions.

**Running time**

We observed that the running times of both algorithms $A3, A4$ depend on two key factors:

- $fdelay$, the time taken per execution of the functional composition at each step

- Complexity of dominance testing which is in turn a function of $|D_i|$, $|\mathcal{X}|$ and the properties of $\{\succ_i\}$ and $\rhd$. In particular, the complexity of dominance testing depends on the size of the preference relations $\{\succ_i\}$ and $\rhd$ (see Section 5.2).

In order to understand the effect of $fdelay$ on the running times of the algorithms, we ran experiments with $fdelay = 10ms$ and $fdelay = 1000ms$ on problem instances where relative importance preferences are interval/total orders and intra-attribute preferences are partial/total orders (see Table 8 for the other parameters used and their ranges). The respective results are shown in Figures 11 – 14. The results yield the following observations.

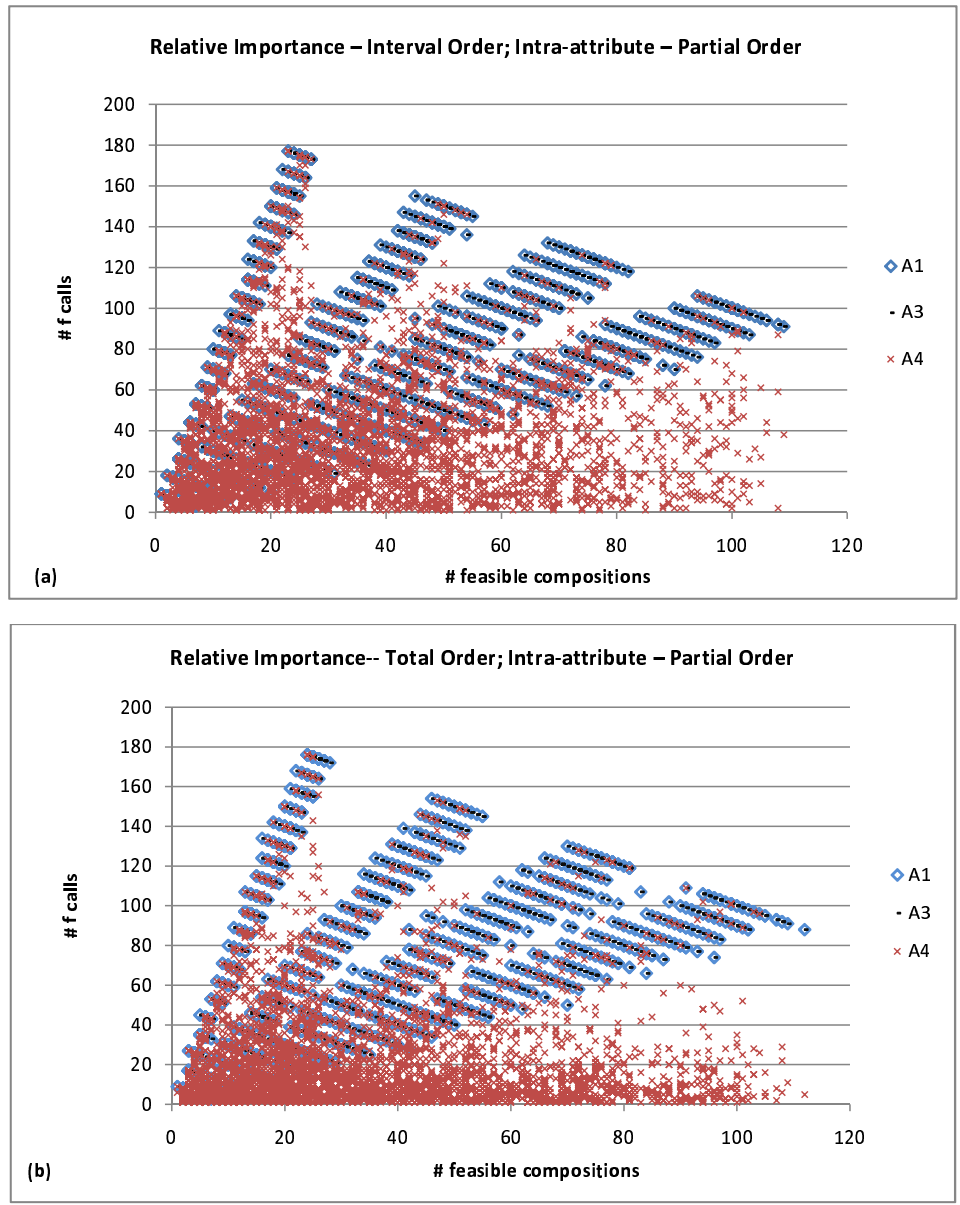

Figure 9: A comparison of the algorithms $A1$, $A3$ and $A4$ with respect to the number of times they invoke the step-by-step functional composition algorithm during their execution. The plots (a) and (b) correspond to results of running the algorithms on simulated problem instances, where the intra-attribute preference ($\succ_i$) is a partial order, and the relative importance preference ($\triangleright$) is an interval or total order. The four distinct "bands" seen in the plots correspond to various fractions of leaves in the search tree of the problem instance that are feasible compositions: $feas = 0.25, 0.5, 0.75, 1.0$.

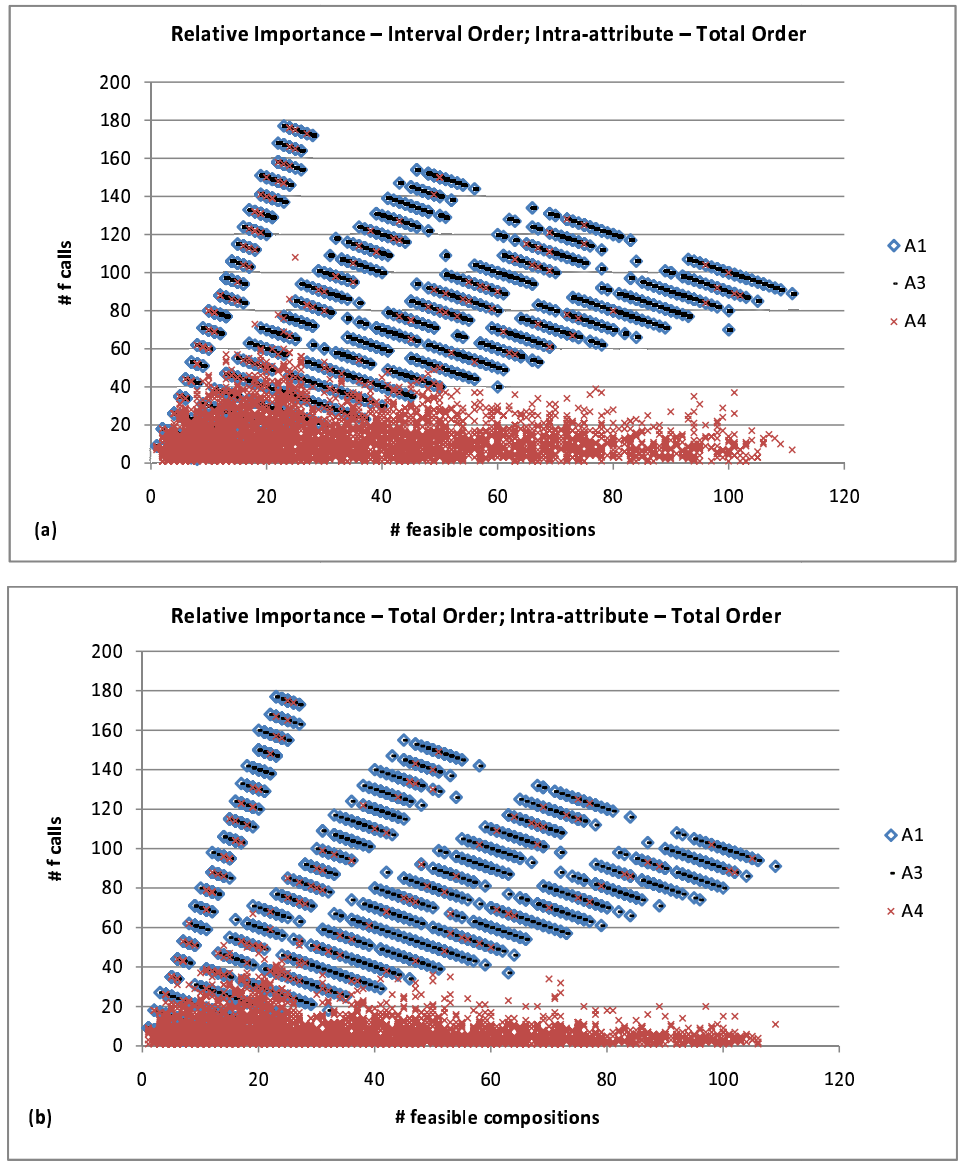

Figure 10: A comparison of the algorithms $A1$, $A3$ and $A4$ with respect to the number of times they invoke the step-by-step functional composition algorithm during their execution. The plots (a) and (b) correspond to results of running the algorithms on simulated problem instances, where the intra-attribute preference ($\succ_i$) is a total order, and the relative importance preference ($\rhd$) is an interval or total order. The four distinct "bands" seen in the plots correspond to various fractions of leaves in the search tree of the problem instance that are feasible compositions: $feas = 0.25, 0.5, 0.75, 1.0$.

- In general, in comparison to the running time of the algorithm $A4$ when the intra-attribute preferences ($\succ_i$) are partial orders, $A4$ is faster when $\succ_i$ are total orders. This trend is observed in plots (a) and (b) of Figure 12 (where intra-attribute preferences are total orders), as the data points corresponding to the running time of $A4$ (colored red) are much closer to the axis corresponding to the number of feasible compositions, in comparison to the plots (a) and (b). A similar trend is also observed in Figures 13 and 14.

- The algorithm $A3$ almost always outperforms the blind search algorithm $A1$ in terms of running time. This is because $A3$ computes the non-dominated set in the last step with respect to the intra-attribute preference over the valuations of one attribute $\succ_i'$ (in place of the dominance relation $\succ_d$ used by $A1$).

- The interleaved algorithm $A4$ is more sensitive to the complexity of dominance than $A1$ and $A3$, because at each step $A4$ computes the non-dominated subset of extensions to explore. On the other hand, $A1$ and $A3$ involve computation of dominance only in the last step. $A3$ is faster than $A1$, more than $A4$, because it computes the non-dominated set with respect to the intra-attribute preference over the valuations of one attribute $\succ_i'$ (in place of the dominance relation $\succ_d$ used by $A1$ and $A4$).

- Algorithms $A1$ and $A3$ are more sensitive to $fdelay$ than the interleaved algorithm $A4$. This is because at each step $A1$ and $A3$ explore all feasible extensions, but $A4$ only explores the preferred subset of the feasible extensions at each step.

- The overall running times of $A1$, $A3$ and $A4$ depend on the relative trade-offs among $|D_i|$, $|\mathcal{X}|$, the properties of $\{\succ_i\}, \triangleright$ (those that influence the complexity of dominance testing) on the one hand and $fdelay$ on the other.

## 7. Summary and Discussion

We now summarize our contributions in this paper.

### 7.1 Summary

Many applications, e.g., planning, Web service composition, embedded system design, etc., rely on methods for identifying collections (compositions) of objects (components) satisfying some functional specification. Among the compositions that satisfy the functional specification (feasible compositions), it is often necessary to identify one or more compositions that are most preferred with respect to user preferences over non-functional attributes. Of particular interest are settings where user preferences over attributes are expressed in qualitative rather than quantitative terms (Doyle & Thomason, 1999).

In this paper, we have proposed a framework for representing and reasoning with qualitative preferences over compositions in terms of the qualitative preferences over attributes of their components; and developed a suite of algorithms to compute the most preferred feasible compositions, given an algorithm that computes the functionally feasible compositions. Specifically,

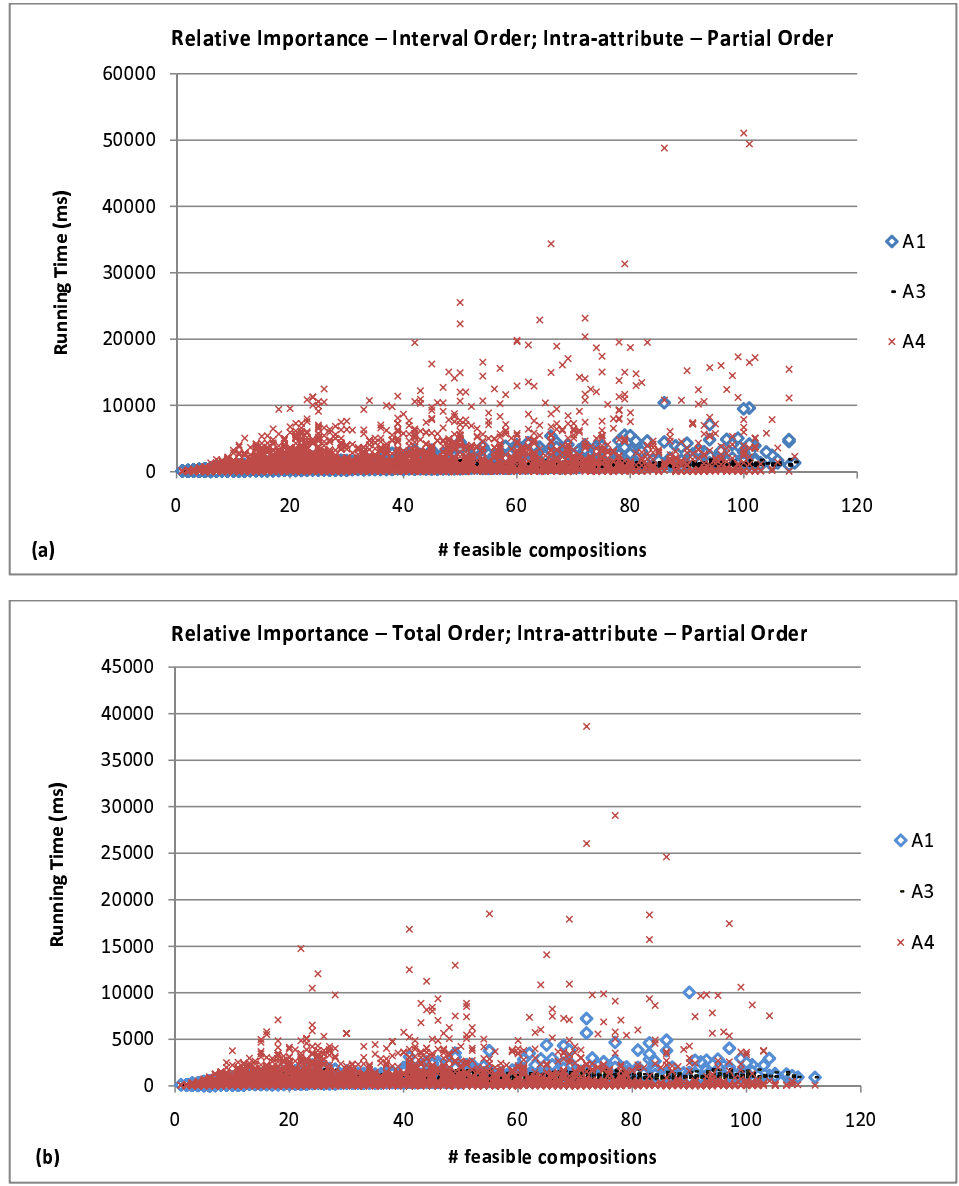

Figure 11: A comparison of the algorithms $A1$, $A3$ and $A4$ with respect to their running times as a function of the number of feasible compositions, when each invocation step in the step-by-step functional composition algorithm has a overhead of 10 milliseconds. The plots (a) and (b) correspond to results of running the algorithms on simulated problem instances, where the intra-attribute preference ($\succ_i$) is a partial order, and the relative importance preference ($\triangleright$) is an interval or total order. The four distinct "bands" seen in the plots correspond to various fractions of leaves in the search tree of the problem instance that are feasible compositions: $feas = 0.25, 0.5, 0.75, 1.0$.

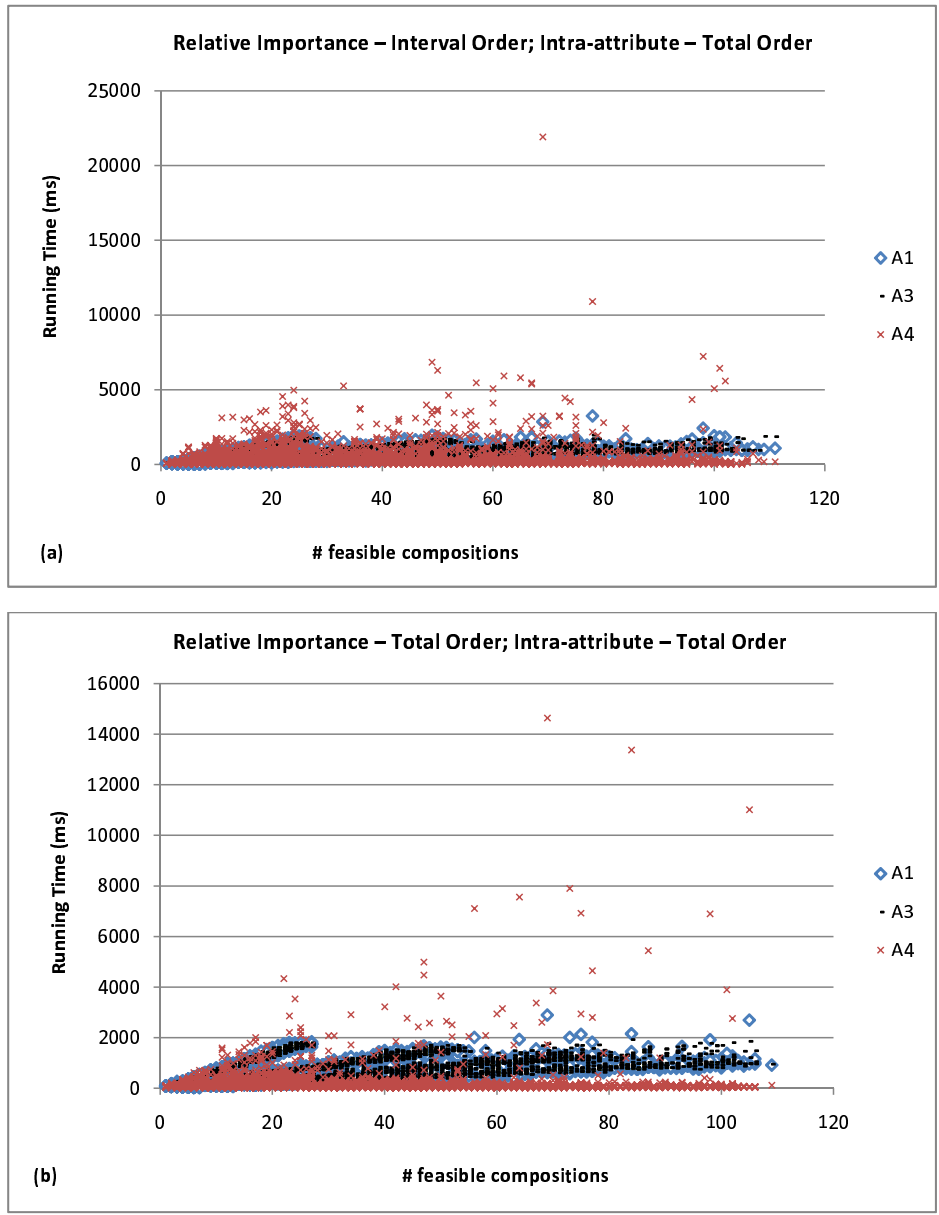

Figure 12: A comparison of the algorithms $A1$, $A3$ and $A4$ with respect to their running times as a function of the number of feasible compositions, when each invocation step in the step-by-step functional composition algorithm has a overhead of 10 milliseconds. The plots (a) and (b) correspond to results of running the algorithms on simulated problem instances, where the intra-attribute preference ($\succ_i$) is a total order, and the relative importance preference ($\rhd$) is an interval or total order. The four distinct "bands" seen in the plots correspond to various fractions of leaves in the search tree of the problem instance that are feasible compositions: $feas = 0.25, 0.5, 0.75, 1.0$.

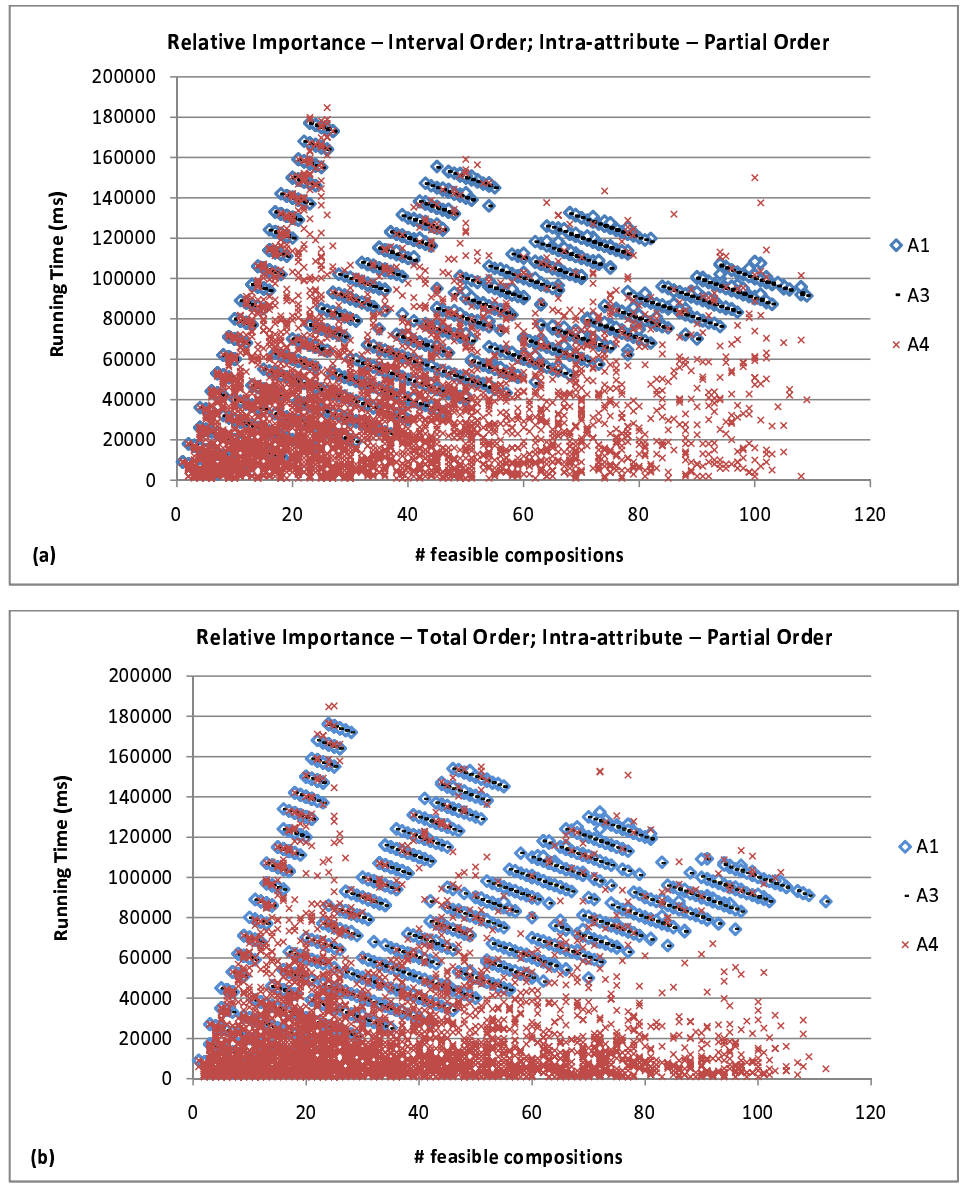

Figure 13: A comparison of the algorithms *A1*, *A3* and *A4* with respect to their running times as a function of the number of feasible compositions, when each invocation step in the step-by-step functional composition algorithm has a overhead of 1000 milliseconds. The plots (a) and (b) correspond to results of running the algorithms on simulated problem instances, where the intra-attribute preference ($\succ_i$) is a partial order, and the relative importance preference ($\rhd$) is an interval or total order. The four distinct "bands" seen in the plots correspond to various fractions of leaves in the search tree of the problem instance that are feasible compositions: $feas = 0.25, 0.5, 0.75, 1.0$.

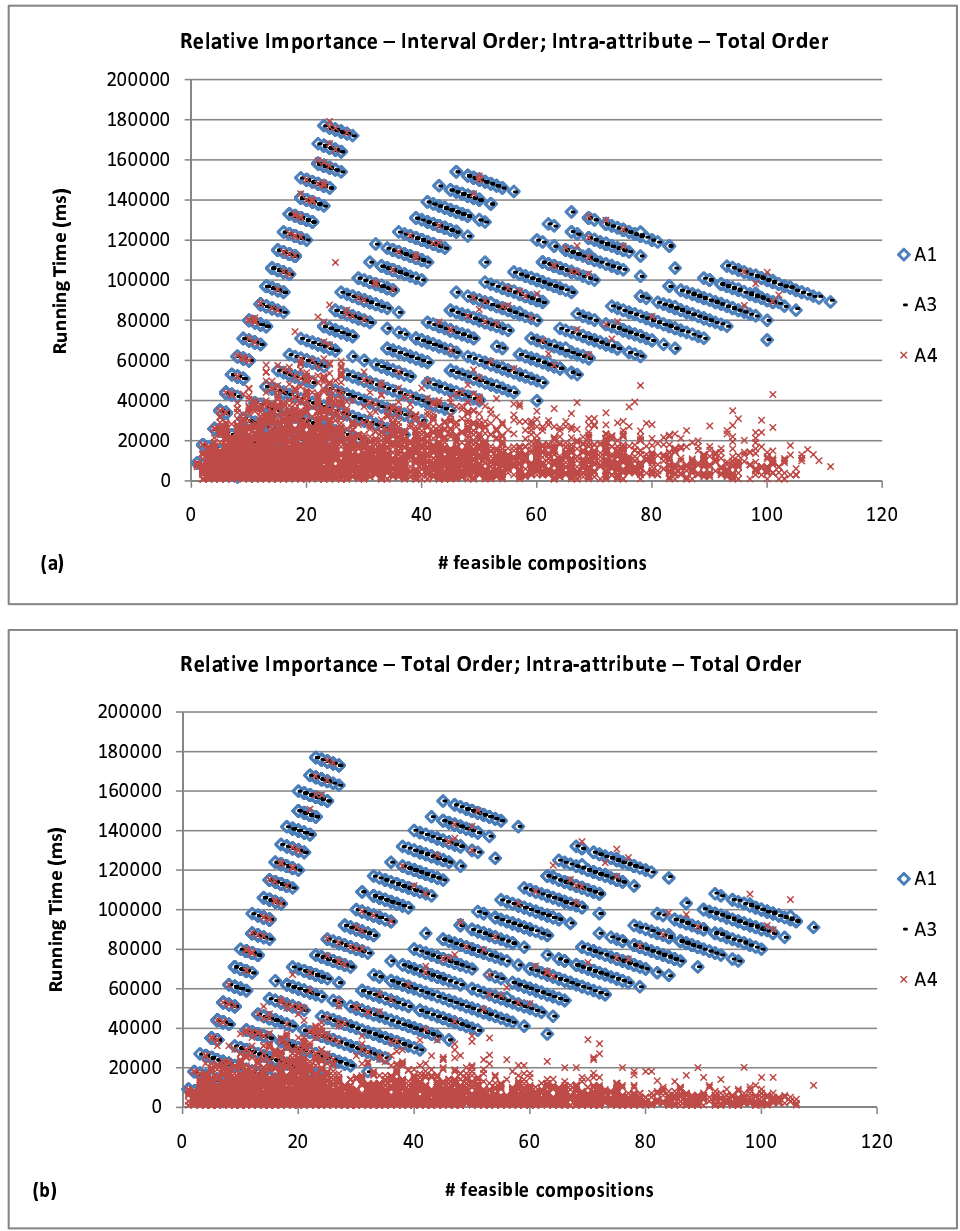

Figure 14: A comparison of the algorithms *A1*, *A3* and *A4* with respect to their running times as a function of the number of feasible compositions, when each invocation step in the step-by-step functional composition algorithm has a overhead of 1000 milliseconds. The plots (a) and (b) correspond to results of running the algorithms on simulated problem instances, where the intra-attribute preference ($\succ_i$) is a total order, and the relative importance preference ($\rhd$) is an interval or total order. The four distinct "bands" seen in the plots correspond to various fractions of leaves in the search tree of the problem instance that are feasible compositions: $feas = 0.25, 0.5, 0.75, 1.0$.

a) We have defined a generic *aggregation function* to compute the valuation of a composition as a function of the valuations of its components. We have also presented a strict partial order preference relation for comparing two compositions with respect to their aggregated valuations of each attribute;

b) We have introduced a *dominance* relation for comparing compositions based on user specified preferences and established some of its key properties. In particular, we have shown that this dominance relation is a strict partial order when intra-attribute preferences are strict partial orders and relative importance preferences are interval orders.

c) We have developed four algorithms for identifying the most preferred composition(s) with respect to the user preferences. The first three algorithms first compute the set of all feasible compositions (solutions) using a functional composition algorithm as a *black box*, and then proceed to find the most preferred among them (1) based on the dominance relation (*ComposeAndFilter*); and (2) based on the preferred valuations with respect to the most important attribute(s) (*WeaklyCompleteCompose* and *AttWeaklyCompleteCompose*). The fourth algorithm interleaves the execution of a functional composition algorithm that produces the set of solutions by iteratively extending partial solutions and the ordering of partial solutions with respect to user preferences (*InterleaveCompose*).

d) We have established some key properties of the above algorithms. *ComposeAndFilter* is guaranteed to return the set of all non-dominated solutions; *WeaklyCompleteCompose* is guaranteed to return a non-empty subset of non-dominated solutions; *AttWeaklyCompleteCompose* is guaranteed to return at least one of the non-dominated solutions; and *InterleaveCompose* is guaranteed to return (i) a non-empty subset of non-dominated solutions when the dominance relation is an interval order; and (ii) the entire set of non-dominated solutions when the dominance relation is a weak order.

e) We have performed simulation experiments to compare the algorithms with respect to (i) the ratio of most preferred solutions produced to the actual set of most preferred solutions, and the ratio of the most preferred solutions produced to the entire set of solutions produced by the algorithm; (ii) their running times as a function of the search space and the overhead in each call to the functional composition algorithm; and (iii) the number of calls each algorithm makes to the functional composition algorithm during the course of its execution. The results showed the feasibility of our algorithms for composition problems that involve up to 200 components.

f) We have analyzed the results of experiments to obtain additional theoretical properties of the dominance relation as a function of the properties of the underlying intra-attribute preference relations and relative importance preference relation. In particular, we obtained non-trivial results as a consequence of our analysis of experimental results, which were not known apriori, including conditions under which the dominance relation is a weak order. These conjectures/results are significant because they give the properties of the dominance relation directly as a function of the input user preferences. In turn, they also throw light on the soundness, weak-completeness and/or completeness properties of the algorithms.

The proposed techniques for reasoning with preferences over non-functional attributes are independent of the language used to express the desired functionality $\varphi$ of the composition, and the method used to check whether a composition $\mathcal{C}$ satisfies the desired functionality, i.e., $\mathcal{C} \models \varphi$. Our formalism and algorithms may be applicable to a broad range of domains including Web service composition (see Dustdar & Schreiner, 2005; Pathak, Basu, & Honavar, 2008, for surveys), planning (see Hendler, Tate, & Drummond, 1990; Baier & McIlraith, 2008a), team formation (see Lappas, Liu, & Terzi, 2009; Donsbach, Tannenbaum, Alliger, Mathieu, Salas, Goodwin, & Metcalf, 2009) and indeed any setting that calls for choosing the most preferred solutions from a set of candidate solutions, where each solution is made up of multiple components.

## 7.2 Discussion

In the following, we discuss some of the alternate choices that one could make in applying our formalism for specific applications.

*Aggregation Functions.* In our previous work (Santhanam, Basu, & Honavar, 2008), we had proposed the use of TCP-net representation with ceteris paribus semantics (Brafman et al., 2006) for reasoning with preferences in addressing the problem of Web service composition. We had assumed that the intra-attribute preferences are total orders; however, this assumption does not hold in many practical settings involving qualitative preferences over non-functional attributes. In this paper, we have relaxed this requirement, allowing intra-attribute preferences that are strict partial orders.

In this paper we demonstrated the use of the summation (e.g., number of credits in a POS) and worst frontier (e.g., areas of study and instructors) aggregation functions. In some scenarios, it might be necessary to consider other ways of aggregating the valuations of the components, for example, using the *best frontier* denoting the best possible valuations of the components (i.e., the maximal valuations for each attribute $X_i$ with respect to $\succ_i$). Any aggregation function can be used in our formalism, provided that the preference relation over the aggregated valuations is a strict partial order. Otherwise, the choice of aggregation function and the preference relation to compare aggregated valuations may impact the properties of the resulting dominance relation, and as a result, may also affect the soundness and completeness properties of some of the proposed algorithms.

The aggregation functions demonstrated in this paper are independent of the *how* the components interact or are assembled, i.e., the *structure* of a composition. However, in general, it may be necessary for the aggregation function to take into account the structure and/or other interactions between the valuations of components in a composition. For example, in evaluating the reliability of a composition, one needs to consider the precise structure of the composition. The reliability of a composition $\mathcal{C}_i$ is the product of the reliabilities of the components ($\prod_{i=1}^{n} V_{W_i}(Reliability)$) when the components are arranged in a series configuration (Rausand & Høyland, 2003). On the other hand, when the same set of components $\{W_i\}$ are arranged in a parallel configuration, the reliability of $\mathcal{C}_i$ is computed as $(1 - \prod_{i=1}^{n}(1 - V_{W_i}(Reliability)))$. In general, it might be necessary to introduce aggregation functions that take into consideration a variety of factors including the structure, the function, as well as the non-functional attributes of the composition.

*Comparing Sets of Aggregated Valuations.* In this paper, we presented a preference relation ($\succ_i'$) to compare sets of valuations computed using the worst frontier aggregation function (Definition 8). This preference relation requires that given two sets of valuations, every element in the dominated set is preferred to at least one of the elements in the dominating set of valuations. Other choices of $\succ_i'$ can be used as well, but care should be taken because the properties of the chosen preference relation may affect the properties of the dominance ($\succ_d$) relation and the properties of the algorithms. However, as long as $\succ_i'$ is a strict partial order (irreflexive and transitive), the dominance relation continues to remain a strict partial order (subject to $\rhd$ being an interval order), and hence the properties of the algorithms hold. This provides the user with a wide range of preference relations for comparing sets of valuations to choose from (see Barbera et al., 2004, for a survey of preferences over sets).

Note that Definition 8 does not ignore common elements when comparing two sets of elements. However, some settings may require a preference relation that compares only elements in the two sets that are not common. In such settings, a suitable irreflexive and transitive preference relation can be used, such as the asymmetric part of preference relations developed by Brewka et al. (2010) and Bouveret et al. (2009). In the absence of transitivity, the transitive closure of the relation may be used to compare sets of elements, as done by Brewka et al.

*Dominance and its Properties.* The dominance relation ($\succ_d$) adopted in this paper is a strict partial order when the intra-attribute preferences are arbitrary strict partial orders and the relative importance is an interval order. It would be interesting to explore alternative notions of dominance that preserve the rationality of choice, by requiring a different set of properties (e.g., those that satisfy *negative-transitivity* instead of transitivity). It would also be of interest to examine the relationships between $\succ_d$ and alternative dominance relations. Some results comparing $\succ_d$ with the dominance relations proposed by other authors (Brafman et al., 2006; Wilson, 2004b, 2004a) have been presented elsewhere (Santhanam, Basu, & Honavar, 2010b, 2009).

*Implementation.* The current implementation of dominance testing with respect to $\succ_d$ is based on iteratively searching all the attributes to find a witness. It would be interesting to compare this with other methods for dominance testing such as the one proposed in one of our earlier works (Santhanam, Basu, & Honavar, 2010a) that uses efficient model checking techniques. We would also like to use other multi-attribute preference formalisms that include conditional preferences in our framework for compositional systems and compare the performance of the resulting implementation with the current implementation.

## 7.3 Related Work

Techniques for representing and reasoning with user preferences over a set of alternatives have been studied extensively in the areas of decision theory, microeconomics, psychol-ogy, operations research, etc. The seminal work by von Neumann and Morgenstern (1944) models user preferences using *utility functions* that map the set of possible alternatives to numeric values. More recently, models for representing and reasoning with quantitative preferences over multiple attributes have been developed (Fishburn, 1970a; Keeney & Raiffa, 1993; Bacchus & Grove, 1995; Boutilier, Bacchus, & Brafman, 2001). Such models have been used to address problems such as identifying the most preferred tuples resulting from database queries (Agrawal & Wimmers, 2000; Hristidis & Papakonstantinou, 2004; Börzsönyi, Kossmann, & Stocker, 2001), assembling preferred composite Web services (Zeng, Benatallah, Dumas, Kalagnanam, & Sheng, 2003; Zeng, Benatallah, Ngu, Dumas, Kalagnanam, & Chang, 2004; Yu & Lin, 2005; Berbner, Spahn, Repp, Heckmann, & Steinmetz, 2006), and in other composition problems.

However, in many applications it is more natural for users to express preferences in qualitative terms (Doyle & McGeachie, 2003; Doyle & Thomason, 1999; Dubois, Fargier, Prade, & Perny, 2002) and hence, there is a growing interest in AI on formalisms for representing and reasoning with qualitative preferences (Brafman & Domshlak, 2009). We now proceed to place our work in the context of some of the recent work on representing and reasoning with qualitative preferences.

### 7.3.1 TCP-NETS

Notable among *qualitative* frameworks for preferences are *preference networks* (Boutilier et al., 2004; Brafman et al., 2006) that deal with qualitative and conditional preferences. A class of preference networks, namely Tradeoff-enhanced Conditional Preference networks (TCP-nets) (Brafman et al., 2006) are closely related to our work, and we now proceed to discuss where our framework departs from and adds to the existing TCP-net framework.

TCP-nets provide a very elegant and compact graphical model to represent *qualitative* intra-attribute and relative importance preferences over a set of attributes. In addition, TCP-nets can also model conditional preferences using dependencies among attributes. While TCP-nets allow us to represent and reason about preferences in general over simple objects (each of which is described by a set of attributes), the focus of our work is to reason about such preferences over *compositions* of simple objects (i.e., a collection of objects satisfying certain functional properties). For example, in the domain of Web services, the problem of identifying the most preferred Web services from a repository of available ones based on their non-functional attributes, namely *Web service selection* can be solved using the TCP-net formalism. On the other hand, in addition to Web service selection, our formalism can also address the more complicated problem of identifying the most preferred composite Web services that collectively satisfy a certain functional requirement, namely *Web service composition.*

Our formalism is based on the intra-attribute and relative importance preferences over a set of attributes describing the objects. As a result, the graphical representation scheme of TCP-nets can still be used to compactly encode the intra-attribute and relative importance preferences of the users within our formalism [14].

---

14. In our setting, we do not consider conditional preferences that correspond to edges denoting conditional dependencies in the TCP-nets.

We have extended reasoning about preferences over single objects to enable reasoning about preferences over collections of objects. We have: (a) provided an aggregation function for computing the valuation of a composition as a function of the valuations of its components; (b) defined a dominance relation for comparing the valuations of compositions and established some of its properties; and (c) developed algorithms for identifying a subset or the set of most preferred composition(s) with respect to this dominance relation.

Our formalism departs from TCP-nets in the interpretation of the intra-attribute and relative importance preferences over objects: the dominance relation in a TCP-net is defined as *any* partial order relation that is *consistent* with the given preferences over attributes of the objects, based on the *ceteris-paribus* semantics. We introduce a dominance relation (see Definition 11) that allows us to reason about preferences over *collections* of objects in terms of *sets* of valuations of the attributes of objects that make up the collection. For instance, our worst frontier aggregation function returns the set of worst possible attribute valuations among all the components.

When our dominance relation is applied in the simpler setting where each collection consists of a *single object*, the aggregation function for each attribute reduces to the identity function, and the preference relation $\succ_i'$ over *sets* of valuations of each attribute $X_i$ reduces to the intra-attribute preference $\succ_i$. We have recently shown in our earlier works (Santhanam et al., 2010b, 2009) that in general, when TCP-nets are restricted to unconditional preferences, our dominance relation (when each collection consists of a single object) and the dominance relation used in TCP-nets are incomparable; when relative importance is restricted to be an interval order, our dominance relation is more general than the dominance relation used in TCP-nets with only unconditional preferences. In the latter case, our dominance relation is computable in polynomial time, whereas there are no known polynomial time algorithms for computing TCP-net dominance (Santhanam et al., 2010b, 2009).

### 7.3.2 Preferences over Collections of Objects

Several authors have considered ways to extend user preferences to obtain a ranking of collections of objects (see Barbera et al., 2004, for a survey). In all these works, preferences are specified over individual objects in a set as opposed to preferences over valuations of the attributes of the objects. The preferences over objects are in turn used to reason about preferences over collections of those objects. This scenario can be simulated by our framework, by introducing a single attribute whose valuations correspond to objects in the domain.

DesJardins et al. (2005) have considered the problem of finding subsets that are optimal with respect to user specified *quantitative* preferences over a set of attributes in terms of the desired *depth*, *feature weight* and *diversity* for each attribute. In contrast, our framework focuses on *qualitative* preferences. In our setting, depth preferences that map attribute valuations to their relative desirability can be mapped to qualitative intra-attribute preferences and *feature weights* can be mapped to relative importance. Diversity preferences over attributes refer to the *spread* (e.g., variance, range, etc.) of component valuations with respect to the corresponding attributes. It would be interesting to explore whether a suitable dominance relation can be defined so as to simultaneously capture in our framework the user preferences with respect to the depth, diversity and feature weights.

| Property | Denoted by | New Attribute | Attribute Domain |
|---|---|---|---|
| Party Affiliation | P | $X_P$ | $\{Re, De\}$ |
| Views | V | $X_V$ | $\{Li, Co, Ul\}$ |
| Experience | E | $X_E$ | $\{Ex, In\}$ |

Table 13: Properties/Attributes describing the senators

More recently, Binshtok et al. (2009) have presented a language for specification of preferences over sets of objects. This framework, in addition to intra-attribute and relative importance preferences over attributes, allows users to express preferences over the *number* ($|\varphi|$) of elements in a set that satisfy a desired property $\varphi$. The preference language in this case allows statements such as "$S_i : |\varphi|\ REL\ n$" (number of elements in the preferred set with property $\varphi$ should be $REL$ $n$), "$S_j : |\varphi|\ REL\ |\psi|$" (number of elements in the preferred set with property $\varphi$ should be $REL$ number of elements in the preferred set with property $\psi$), etc., where $REL$ is one of the arithmetic operators $>, <, =, \geq, \leq$ and $n$ is an integer. In addition, there can be relative importance between the various preference statements such as "$S_i$ is more important than $S_j$" as well as *external* cardinality constraints such as a bound on the number of elements in the preferred set.

Our formalism can accommodate such preference statements, by representing each preference statement $S_i$ as a new binary valued attribute in the compositional system. For example, preference statements $S_i : |\varphi| \geq n$ and $S_j : |\varphi| \leq |\psi|$ can be represented in our formalism by creating new binary attributes $X_i$ and $X_j$ with intra-attribute preferences $1 \succ_i 0$ and $1 \succ_j 0$ respectively. The relative importance statements such as "$S_i$ is more important than $S_j$" can then be directly mapped to $X_i \rhd X_j$. Any external cardinality constraints on the size of the preferred set can be encoded in our setting by *functional* requirements, so as to restrict the feasible solutions to only those that satisfy the cardinality constraints.

Consider the example discussed by Binshtok et al. (2009), with preferences over senate members described by attributes: Party affiliation (Republican, Democrat ), Views (liberal, conservative, ultra conservative), and Experience (experienced, inexperienced). The attributes and their domains are listed in Table 13. The set preferences are given by:

- $S_1 : \langle |P = Re \lor V = Co| \geq 2 \rangle$

- $S_2 : \langle |E = Ex| \geq 2 \rangle$

- $S_3 : \langle |V = Li| \geq 1 \rangle$

Note that the senate members (i.e., the individual objects) are described by three attributes $X_P, X_V, X_E$ representing the party affiliation, views and experience respectively. The valuation function for these attributes is defined in the obvious manner, e.g., if a senator $W_j$ is a republican, then $V_{W_j}(X_P) = Re$. We introduce three additional boolean attributes $X_1, X_2, X_3$ corresponding to the preference statements $S_1, S_2, S_3$ respectively. The valuation function for each new attribute of a senator $W_i$ can then be defined as follows.

- $V_{W_i}(X_1) = \begin{cases} 1 \text{ , if } W_i \models S_1 \text{ i.e., } V_{W_i}(X_P) = Re \text{ or } V_{W_i}(X_V) = Co \\ 0 \text{ , otherwise} \end{cases}$

- $V_{W_i}(X_2) = \begin{cases} 1 \text{ , if } W_i \models S_2 \text{ i.e., } V_{W_i}(X_E) = Ex \\ 0 \text{ , otherwise} \end{cases}$

- $V_{W_i}(X_3) = \begin{cases} 1 \text{ , if } W_i \models S_3 \text{ i.e., } V_{W_i}(X_V) = Li \\ 0 \text{ , otherwise} \end{cases}$

The valuation of the collection of senators $W_1 \oplus W_2 \oplus \ldots \oplus W_n$ for $i \in \{1, 2, 3\}$ is:

$$V_{W_1 \oplus W_2 \oplus \ldots \oplus W_n}(X_i) = \Phi_i(V_{W_1}, V_{W_2}, \ldots V_{W_n}) = V_{W_1}(X_i) + V_{W_2}(X_i) + \cdots + V_{W_n}(X_i)$$

Note that the aggregation function $\Phi_i$ defined above differs from the worst-frontier based aggregation function adopted in Definition 6. The preference relation for comparing groups of senators with respect to each new attribute $X_i$ can then be defined based on the preference statement $S_i$. For example, in the case of $X_1$ we define $\succ'_1$ such that any value $\geq 2$ is preferred to any value $< 2$, etc. Having defined the above aggregation function and comparison relation for each new attribute, any dominance relation can be adopted to compare compositions (arbitrary subsets) with respect to all attributes including the dominance relation used by Binshtok et al. (2009).

In contrast to the framework of Binshtok et al., (2009) our formalism focuses on collections of objects that satisfy some desired criteria, rather than arbitrary subsets. We provide algorithms for finding the most preferred compositions that satisfy the desired criteria.

### 7.3.3 Database Preference Queries

Several authors (Börzsönyi et al., 2001; Chomicki, 2003; Kiessling & Kostler, 2002; Kiessling, 2002) have explored techniques for incorporating user specified preferences over the result sets of relational database queries. For instance, Chomicki's framework (2003) allows user preferences over each of the attributes of a relation to be expressed as first order logic formulas. Suppose $S_q$ is the set of tuples that match a query $q$. For each attribute $X_i$, from $S_q$, a subset $S_{q_i}$ of tuples that have the most preferred value(s) for $X_i$ is identified. The result set for the query $q$ is then given by $\cap_i S_{q_i}$. A similar framework for expressing and combining user preferences is presented by Kiessling (2002) and Kiessling and Kostler (2002). Brafman and Domshlak (2004) have pointed out some of the semantic difficulties associated with above approaches, and considered an alternative approach to identifying the preferred result set based on the CP-net (Boutilier et al., 2004) dominance relation. Because of the high computational complexity of dominance testing for CP-nets, Boutilier et al. proposed an efficient alternative based on an ordering operator that orders the tuples in the result set in a way that is consistent with the user preferences. Our formalism can be used in the database setting, similar in spirit to that of Brafman and Domshlak, by considering each tuple in $S_q$ as a collection with a single object. The differences in the semantics of the CP-net dominance and our dominance relation is discussed in Section 7.3.1.

A host of algorithms have also been proposed for computing the non-dominated result set in response to preference queries, especially for the efficient evaluation of *skyline* queries (Börzsönyi et al., 2001; Chomicki, 2003). A skyline query yields the non-dominated result set from a database, where dominance is evaluated based on the notion of *pareto dominance* that considers all attributes to be equally important. Most of the proposed algorithms for computing the skylines (see Jain, 2009, for a survey) are applicable only when intra-attribute preferences are totally or weakly ordered. Some other algorithms that can handle partially ordered attribute domains (Chan, Eng, & Tan, 2005; Sacharidis, Papadopoulos, & Papadias, 2009; Jung, Han, Yeom, & Kang, 2010) rely on creating and maintaining indexes over the attributes in the database, and on data structures specifically designed to identify the skyline with respect to pareto dominance. These algorithms may be considered if a particular problem instance involves such a large set of components are already stored in a database and indexed. However, it is not obvious that they generalize to an arbitrary notion of dominance such as the one presented here. On the other hand, our algorithms for finding the non-dominated set are applicable to any notion of dominance, provided the user preferences are such that the dominance relation is a partial order.

### 7.3.4 PLANNING WITH PREFERENCES

The classical planning problem consists of finding a sequence of actions that take a system from an initial state to one of the states that satisfies the user specified goal. Preference based planning refers to the problem of finding plans that are most preferred with respect to a set of user preferences over the plans. Such preferences are usually compactly expressed in terms of the preferences over the properties satisfied by the plans in the goal or intermediate states, or over actions, or over action sequences (i.e., temporal properties of the plans). We refer the interested reader to surveys by Baier et al. (2008b) and Bienvenu et al. (2011) for an overview of qualitative and quantitative preference languages used in preference based AI planning, and different algorithms for computing the most preferred plans.

Preference based planning can be viewed as a problem of finding the most preferred composition in a compositional system, where the components correspond to the actions, and the feasible compositions correspond to the states of the plans that satisfy the goal in the planning problem. The allowed set of actions that can be performed from a given state in the planning problem can be encoded in the compositional system in terms of a set of functional requirements (or constraints on the functionality). The preferences over the various actions that can be taken at any given state in a plan can be captured by preferences over the components with which a composition can be extended in terms of their properties or attribute valuations. The properties satisfied by a state of a plan in the planning problem can be captured by the valuations of the attributes of the corresponding composition in the compositional system. Based on the mapping of actions performed in a given state to the properties of the resulting state in the planning problem, aggregation functions can be suitably defined in the compositional system. The addition of an action to a partial plan in the planning problem can be represented in the compositional system by the extension of a partial composition by a new component, and the properties satisfied by the resulting state in the planning problem correspond to the valuations of the attributes of the extended composition as determined by the aggregation functions. Finding the most preferred plans then involves finding the most preferred feasible compositions.

The algorithms presented in this paper can be used to find the most preferred plans with respect to the user specified preferences over actions in terms of the properties satisfied by their resulting states, or over the properties satisfied by the plans in the goal state. However, planning problems that involve preferences over the orderings of states and actions in a plan, e.g., preferences over the properties that hold over the entire sequence of states of a plan (Baier, Bacchus, & McIlraith, 2009; Bienvenu et al., 2011) cannot be handled within our framework.

## 8. Acknowledgments

Aspects of this work were supported in part by NSF grants CNS0709217, CCF0702758, IIS0711356 and CCF1143734. The work of Vasant Honavar was supported by the National Science Foundation, while working at the Foundation. Any opinion, finding, and conclusions contained in this article are those of the authors and do not necessarily reflect the views of the National Science Foundation.

We are grateful to anonymous reviewers for a thorough review and Dr. Ronen Brafman for many useful suggestions that have helped improve the manuscript.

## Appendix A. Proofs of Propositions and Theorems in Section 4

**Proposition 15** $\forall X_i \in I : \Psi_{\succ_d}(\mathfrak{C}) \neq \emptyset \Rightarrow \Psi_{\succ'_i}(\mathfrak{C}) \cap \Psi_{\succ_d}(\mathfrak{C}) \neq \emptyset$.

*Proof.* Let $X_i \in I$ and $\mathcal{U} \in \Psi_{\succ'_i}(\mathfrak{C})$. There are two possibilities: $\mathcal{U} \in \Psi_{\succ_d}(\mathfrak{C})$ and $\mathcal{U} \notin \Psi_{\succ_d}(\mathfrak{C})$. If $\mathcal{U} \in \Psi_{\succ_d}(\mathfrak{C})$, then there is nothing left to prove.

Suppose that $\mathcal{U} \notin \Psi_{\succ_d}(\mathfrak{C})$. Then we show that $\exists \mathcal{V} \neq \mathcal{U}$ such that $\mathcal{V} \in \Psi_{\succ'_i}(\mathfrak{C}) \cap \Psi_{\succ_d}(\mathfrak{C})$.

$$\mathcal{U} \in \Psi_{\succ'_i}(\mathfrak{C}) \wedge \mathcal{U} \notin \Psi_{\succ_d}(\mathfrak{C}) \Rightarrow \exists \mathcal{V} \in \Psi_{\succ_d}(\mathfrak{C}) : \mathcal{V} \succ_d \mathcal{U}.$$

By Definitions 11 and 16, it follows that $\nexists \mathcal{V} \in \Psi_{\succ_d}(\mathfrak{C}) : \mathcal{V}(X_i) \succ'_i \mathcal{U}(X_i)$. Hence, $X_i$ cannot be a witness for $\mathcal{V} \succ_d \mathcal{U}$. Now there are two cases to consider.

*Case 1:* $\mathcal{U}(X_i) \succ'_i \mathcal{V}(X_i)$.

Let attribute $X_j \neq X_i$ be a witness for $\mathcal{V} \succ_d \mathcal{U}$. Since $X_i \in I$, $(X_i \rhd X_j) \vee (X_i \sim_{\rhd} X_j)$. It therefore follows that $\mathcal{V}(X_i) \succeq'_i \mathcal{U}(X_i)$, which contradicts our assumption that $\mathcal{U}(X_i) \succ'_i \mathcal{V}(X_i)$. Hence, $\mathcal{U}(X_i) \not\succ'_i \mathcal{V}(X_i)$.

*Case 2:* $\mathcal{U}(X_i) \sim'_i \mathcal{V}(X_i)$.

Let attribute $X_j \neq X_i$ be a witness for $\mathcal{V} \succ_d \mathcal{U}$. Since $X_i \in I$, $(X_i \rhd X_j) \vee (X_i \sim_{\rhd} X_j)$. From Definition 11, $\mathcal{V} \succ_d \mathcal{U}$ only if $\mathcal{V}(X_i) \succeq'_i \mathcal{U}(X_i)$. Because of our assumption that $\mathcal{U}(X_i) \sim'_i \mathcal{V}(X_i)$, it must be the case that $\mathcal{V}(X_i) = \mathcal{U}(X_i)$, i.e., $\mathcal{V} \in \Psi_{\succ'_i}(\mathfrak{C})$. Thus, we have:

$$\mathcal{U} \in \Psi_{\succ'_i}(\mathfrak{C}) \setminus \Psi_{\succ_d}(\mathfrak{C}) \Rightarrow \exists \mathcal{V} \in \Psi_{\succ'_i}(\mathfrak{C}) \cap \Psi_{\succ_d}(\mathfrak{C}) : \mathcal{V} \succ_d \mathcal{U} \tag{5}$$

This completes the proof. □

**Theorem 4** [Soundness and Weak Completeness of Algorithm 2] Given a set of attributes $\mathcal{X}$, preference relations $\rhd$ and $\succ'_i$, Algorithm 2 generates a set $\theta$ of feasible compositions such that $\theta \subseteq \Psi_{\succ_d}(\mathfrak{C})$ and $\Psi_{\succ_d}(\mathfrak{C}) \neq \emptyset \Rightarrow \theta \neq \emptyset$.

*Proof.*
*Soundness:* The proof proceeds by contradiction. Suppose that the algorithm returns a solution $\mathcal{U} \in \theta$ such that $\mathcal{U} \notin \Psi_{\succ_d}(\mathfrak{C})$. Because $\mathcal{U} \in \theta$, it is necessary (by Line 5) that $\exists X_i \in I : \mathcal{U} \in \Psi_{\succ'_i}(\mathfrak{C}) \setminus \Psi_{\succ_d}(\mathfrak{C})$. Then, from Equation (5) in the proof of Proposition 15, $\exists \mathcal{V} \in \Psi_{\succ'_i}(\mathfrak{C}) \cap \Psi_{\succ_d}(\mathfrak{C}) : \mathcal{V} \succ_d \mathcal{U}$, which means that $\mathcal{U} \notin \Psi_{\succ_d}(\Psi_{\succ'_i}(\mathfrak{C}))$. However, this contradicts Line 5 of the algorithm. Hence, $\theta \subseteq \Psi_{\succ_d}(\mathfrak{C})$, i.e., Algorithm 2 is sound.

*Weak Completeness:* Because $I \neq \emptyset$, Line 5 is executed by the algorithm at least once for some $X_i \in I$. By Definition 13, we have $\mathfrak{C} \neq \emptyset \Rightarrow \Psi_{\succ'_i}(\mathfrak{C}) \neq \emptyset \Rightarrow \Psi_{\succ_d}(\Psi_{\succ'_i}(\mathfrak{C})) \neq \emptyset \Rightarrow \theta \neq \emptyset$. Hence, Algorithm 2 is weakly complete by Definition 15. $\qquad \blacksquare$

**Proposition 16** If $I = \{X_t\} \land \forall X_k \neq X_t \in \mathcal{X} : X_t \rhd X_k$, then $\Psi_{\succ_d}(\mathfrak{C}) \subseteq \theta$, i.e., Algorithm 2 is complete.

*Proof.* The proof proceeds by contradiction. Let $I = \{X_t\}$ and $\forall X_k \neq X_t \in \mathcal{X} : X_t \rhd X_k$, and suppose that $\exists \mathcal{V} \in \Psi_{\succ_d}(\mathfrak{C}) \setminus \Psi_{\succ'_t}(\mathfrak{C})$. Since $\mathcal{V} \notin \Psi_{\succ'_t}(\mathfrak{C})$, by Definition 13 it must be the case that $\exists \mathcal{U} \in \Psi_{\succ'_t}(\mathfrak{C}) : \mathcal{U}(X_t) \succ'_t \mathcal{V}(X_t)$. However, then $\mathcal{U} \succ_d \mathcal{V}$ by Definition 11 thus contradicting our assumption that $\mathcal{V} \in \Psi_{\succ_d}(\mathfrak{C})$. $\qquad \blacksquare$

**Proposition 17** If $|I| = 1$, i.e., there is a unique most important attribute with respect to $\rhd$, then Algorithm 3 is complete.

*Proof.* Let $I = \{X_i\}$. We know from Proposition 14 that $\succ'_i \subseteq \succ_d$. It follows that $\Psi_{\succ_d}(S) \subseteq \Psi_{\succ'_i}(S)$ for any set $S$. Hence, $\Psi_{\succ_d}(\mathfrak{C}) \subseteq \Psi_{\succ'_i}(\mathfrak{C}) = \theta$, i.e., Algorithm 3 is complete. $\qquad \blacksquare$

**Proposition 18** [Termination of Algorithm 4] Given a finite repository of components, Algorithm 4 terminates in a finite number of steps.

*Proof.* Given a finite repository $R$ of components, and an algorithm $f$ that computes feasible extensions of partial feasible compositions[15], and due to the fact that Algorithm 4 does not re-visit any partial feasible composition, the number of recursive calls is finite. $\qquad \blacksquare$

**Proposition 19** [Unsoundness of Algorithm 4] Given a functional composition algorithm $f$ and user preferences $\succ'_i$ and $\rhd$ over a set of attributes $\mathcal{X}$, Algorithm 4 is not guaranteed to generate a set of feasible compositions $\theta$ such that $\theta \subseteq \Psi_{\succ_d}(\mathfrak{C})$.

*Proof.* We provide an example wherein Algorithm 4 returns a feasible composition that is dominated by some other feasible composition. Consider a compositional system with a single attribute $\mathcal{X} = \{X_1\}$, with a domain of $\{a_1, a_2, a_3, a_4\}$. Let the intra-attribute preference of the user over those values be the partial order: $a_4 \succ_1 a_1$ and $a_2 \succ_1 a_3$ (Figure 15). Let $R = \{W_1, W_2, W_3, W_4\}$ be the repository of components in the compositional system such that $V_{W_i}(X_1) = \{a_i\}$.

---

15. An $f$ that terminates with a set of feasible extensions is guaranteed by the decidability of $\varphi$.

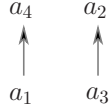

Figure 15: Intra-attribute preference $\succ_1$ for attribute $X_1$

Suppose that there are three feasible compositions in $\mathfrak{C}$ satisfying the user specified functionality $\varphi$, namely $\mathcal{C}_1 = W_1, \mathcal{C}_2 = W_2, \mathcal{C}_3 = W_3 \oplus W_4$. Their respective valuations are: $V_{\mathcal{C}_1} = \langle \{a_1\} \rangle$, $V_{\mathcal{C}_2} = \langle \{a_2\} \rangle$ and $V_{\mathcal{C}_3} = \langle \{a_3, a_4\} \rangle$. Clearly, $\Psi_{\succ_d}(\mathfrak{C}) = \{\mathcal{C}_2, \mathcal{C}_3\}$, because $V_{\mathcal{C}_3} \succ_d V_{\mathcal{C}_1}$ (due to the fact that $\{a_3, a_4\} \succ'_1 \{a_1\}$).

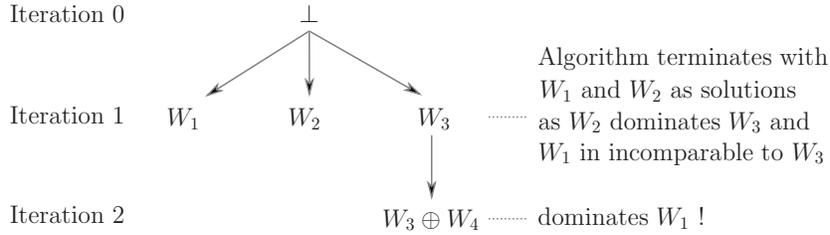

Figure 16: Execution of Algorithm 4

Now suppose that there exists a functional composition algorithm $f$ that produces the following sequence of partial feasible compositions (Figure 16): $\{\perp\}$, $\{W_1, W_2, W_3\}$, $\{W_1, W_2, W_3 \oplus W_4\}$. According to Line 13 of Algorithm 4, the algorithm will terminate after the first invocation of $f$, i.e., when the set $\{W_1, W_2, W_3\}$ of partial feasible compositions is produced by $f$. This is because after the first iteration, $\theta = \{W_1, W_2\}$, with $V_{W_2} \succ_d V_{W_3}$, and both $W_1$ and $W_2$ are feasible compositions. This results in $\theta = \{\mathcal{C}_1, \mathcal{C}_2\} \not\subseteq \Psi_{\succ_d}(\mathfrak{C})$. $\square$

**Theorem 5** [Soundness of Algorithm 4] If $\succ_d$ is an interval order, then given a functional composition algorithm $f$ and user preferences $\{\succ'_i\}, \triangleright$ over a set of attributes $\mathcal{X}$, Algorithm 4 generates a set $\theta$ of feasible compositions such that $\theta \subseteq \Psi_{\succ_d}(\mathfrak{C})$.

*Proof.* Suppose that by contradiction, $\mathcal{F} \in \theta$ and there is a feasible composition $\mathcal{C} \notin \theta$ such that $V_{\mathcal{C}} \succ_d V_{\mathcal{F}}$. If $\mathcal{C}$ is present in the list $\mathcal{L}$ upon termination of the algorithm, then $\mathcal{C}$ should have been in $\theta$, because the algorithm terminates only when *all* compositions in $\Psi_{\succ_d}(\mathcal{L})$ are feasible. This implies that the algorithm did not terminate with an $\mathcal{L}$ containing $\mathcal{C}$.

The algorithm keeps track of all partial feasible compositions that can be extended from $\perp$ in $\mathcal{L}$, without discarding any of them before termination. Therefore, the existence of any such feasible composition $\mathcal{C}$ that is not in $\mathcal{L}$ at the time of termination must imply the existence of some partial feasible composition $\mathcal{B}$ in the list (at the time of termination) that can be extended to produce the feasible composition $\mathcal{C}$, i.e., $\mathcal{B} \oplus W_1 \oplus W_2 \oplus \ldots \oplus W_n = \mathcal{C}$ such that $\mathcal{B} \not\models \varphi$ and $\mathcal{C} \models \varphi$.

$\mathcal{B} \not\models \varphi \Rightarrow \mathcal{B} \notin \theta$ at the time of termination, and therefore $\exists \mathcal{E} \in \theta : V_{\mathcal{E}} \succ_d V_{\mathcal{B}}$. Because $\succ_d$ is transitive (by Proposition 12), since $V_{\mathcal{C}} \not\succ_d V_{\mathcal{B}}$ (by Proposition 6), it follows that $V_{\mathcal{C}} \not\succ_d V_{\mathcal{E}}$ (otherwise, $V_{\mathcal{C}} \succ_d V_{\mathcal{E}} \wedge V_{\mathcal{E}} \succ_d V_{\mathcal{B}} \Rightarrow V_{\mathcal{C}} \succ_d V_{\mathcal{B}}$, a contradiction). Hence, $\mathcal{C}$ must

269

dominate some composition other than $\mathcal{E}$, say $\mathcal{F} \in \theta$ at the time of termination, i.e., $V_{\mathcal{C}} \succ_d V_{\mathcal{F}}$. Because $\mathcal{E}, \mathcal{F} \in \theta$, it follows that $V_{\mathcal{F}} \sim_d V_{\mathcal{E}}$, which in turn implies that $V_{\mathcal{E}} \not\succ_d V_{\mathcal{C}}$. Therefore, $\exists \mathcal{F} \in \theta : V_{\mathcal{C}} \succ_d V_{\mathcal{F}}$, $V_{\mathcal{F}} \sim_d V_{\mathcal{E}}$ and $V_{\mathcal{E}} \sim_d V_{\mathcal{C}}$ (see Figure 17).

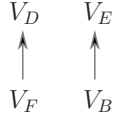

Figure 17: Dominance relationships that violate the interval order restriction on $\succ_d$

From $V_{\mathcal{E}} \succ_d V_{\mathcal{B}}$, $V_{\mathcal{C}} \succ_d V_{\mathcal{F}}$, $V_{\mathcal{F}} \sim_d V_{\mathcal{E}}$ and $V_{\mathcal{C}} \not\succ_d V_{\mathcal{B}}$, it follows that $V_{\mathcal{C}} \sim_d V_{\mathcal{B}}$ (because $V_{\mathcal{B}} \succ_d V_{\mathcal{C}}$ would otherwise imply $V_{\mathcal{E}} \succ_d V_{\mathcal{F}}$, a contradiction). Finally, it must be the case that: $V_{\mathcal{B}} \not\succ_d V_{\mathcal{F}}$, since otherwise it would contradict $V_{\mathcal{F}} \sim_d V_{\mathcal{E}}$; and $V_{\mathcal{F}} \not\succ_d V_{\mathcal{B}}$, since otherwise it would contradict $V_{\mathcal{C}} \sim_d V_{\mathcal{B}}$. Therefore, $V_{\mathcal{B}} \sim_d V_{\mathcal{F}}$. Thus, the only possible dominance relationships among the compositions $\mathcal{B}, \mathcal{C}, \mathcal{E}, \mathcal{F}$ are as follows (see Figure 17):

- $V_{\mathcal{E}} \succ_d V_{\mathcal{B}}$

- $V_{\mathcal{C}} \succ_d V_{\mathcal{F}}$

However, this scenario is ruled out by the fact that $\succ_d$ is an interval order. Hence $\forall \mathcal{F} \in \theta, \forall \mathcal{C} \in \mathfrak{C} \setminus \theta : V_{\mathcal{C}} \not\succ_d V_{\mathcal{F}}$, i.e., $\theta \subseteq \Psi_{\succ_d}(\mathfrak{C})$. $\qquad\blacksquare$

**Theorem 6** [Weak Completeness of Algorithm 4] If $\succ_d$ is an interval order, then given a functional composition algorithm $f$ and user preferences $\{\succ'_i\}, \triangleright$ over a set of attributes $\mathcal{X}$, Algorithm 4 produces a set $\theta$ of feasible compositions such that $\Psi_{\succ_d}(\mathfrak{C}) \neq \emptyset \Rightarrow \theta \cap \Psi_{\succ_d}(\mathfrak{C}) \neq \emptyset$.

*Proof.* From Theorem 5, we have $\theta \subseteq \Psi_{\succ_d}(\mathfrak{C})$ when $\succ_d$ is an interval order. It suffices to show that $\Psi_{\succ_d}(\mathfrak{C}) \neq \emptyset \Rightarrow \theta \neq \emptyset$. The algorithm terminates with the non-dominated set of compositions in the current list $\mathcal{L}$, i.e., the maximal elements of $\mathcal{L}$ with respect to $\succ_d$. The set of maximal elements of any partial order on the set of elements in $\mathcal{L}$ is not empty whenever $\mathcal{L}$ is not empty, and the set of elements in $\mathcal{L}$ is in turn not empty whenever $\mathfrak{C}$ is not empty. Therefore, $\Psi_{\succ_d}(\mathfrak{C}) \neq \emptyset \Rightarrow \mathfrak{C} \neq \emptyset \Rightarrow \mathcal{L} \neq \emptyset \Rightarrow \theta \neq \emptyset$ as required. $\qquad\blacksquare$

**Theorem 7** [Completeness of Algorithm 4] If $\succ_d$ is a weak order, then given a functional composition algorithm $f$ and user preferences $\{\succ'_i\}, \triangleright$ over a set of attributes $\mathcal{X}$, Algorithm 4 generates a set $\theta$ of feasible compositions such that $\Psi_{\succ_d}(\mathfrak{C}) \subseteq \theta$.

*Proof.* It suffices to show that there is no feasible composition $\mathcal{C} \in \Psi_{\succ_d}(\mathfrak{C}) \setminus \theta$.

Suppose by contradiction that $\mathcal{C} \in \Psi_{\succ_d}(\mathfrak{C})$, and $\mathcal{C} \notin \theta$. This means that $\mathcal{C}$ was not present in the list $\mathcal{L}$ upon the termination of the algorithm (because otherwise $\mathcal{C} \in \theta$ as per Lines $4, 6, 13$ in Algorithm 4). Hence, $\mathcal{C}$ must be a feasible extension of some partial feasible composition $\mathcal{B}$ that is present in $\mathcal{L}$ at the time of termination such that $\mathcal{B} \oplus W_1 \oplus W_2 \oplus \ldots \oplus W_k = \mathcal{C}$.

From Proposition 6, we have $V_{\mathcal{C}} \not\succ_d V_{\mathcal{B}}$. Because $\succ_d$ is a weak order, (a) $\forall \mathcal{E} \in \theta : V_{\mathcal{E}} \succ_d V_{\mathcal{B}}$; and (b) $V_{\mathcal{C}} \not\succ_d V_{\mathcal{B}} \wedge V_{\mathcal{E}} \succ_d V_{\mathcal{B}} \Rightarrow V_{\mathcal{E}} \succ_d V_{\mathcal{C}}$. However, this contradicts our assumption that $\mathcal{C} \in \Psi_{\succ_d}(\mathfrak{C})$. $\qquad\blacksquare$